\documentclass[nohyperref]{article}

\usepackage{microtype}
\usepackage{graphicx}
\usepackage{subfigure}
\usepackage{booktabs} 
\usepackage{listings}
\usepackage{xcolor}
\usepackage{comment}
\usepackage{amsmath,amsfonts,amssymb,amsthm,bbm,bm, braket}
\usepackage{tensor}
\usepackage{tikz}
\usepackage{tikz-network}
\usetikzlibrary{patterns,decorations.pathreplacing}
\usepackage{comment}

\tikzset{
    arrowhead/.pic = {
    \draw[thick, rotate = 45] (0,0) -- (#1,0);
    \draw[thick, rotate = 45] (0,0) -- (0, #1);
    }
}

\usepackage{hyperref}

\usepackage[accepted]{icml2023}

\usepackage{amsmath}
\usepackage{amssymb}
\usepackage{mathtools}
\usepackage{amsthm}

\usepackage[capitalize,noabbrev]{cleveref}

\theoremstyle{plain}
\newtheorem{theorem}{Theorem}[section]

\newtheorem{lemma}[theorem]{Lemma}
\newtheorem{corollary}[theorem]{Corollary}
\theoremstyle{definition}
\newtheorem{definition}[theorem]{Definition}

\theoremstyle{remark}

\usepackage[textsize=tiny]{todonotes}

\icmltitlerunning{Simplex Random Features}

\begin{document}

\twocolumn[
\icmltitle{Simplex Random Features}



\icmlsetsymbol{equal}{*}

\begin{icmlauthorlist}
\icmlauthor{Isaac Reid}{cam}
\icmlauthor{Krzysztof Choromanski}{equal,google,columbia}
\icmlauthor{Valerii Likhosherstov}{cam}
\icmlauthor{Adrian Weller}{equal,cam,turing}
\end{icmlauthorlist}

\icmlaffiliation{cam}{University of Cambridge}
\icmlaffiliation{google}{Google}
\icmlaffiliation{columbia}{Columbia University}
\icmlaffiliation{turing}{Alan Turing Institute}

\icmlcorrespondingauthor{Isaac Reid}{ir337@cam.ac.uk}
\icmlcorrespondingauthor{Krzysztof Choromanski}{kchoro@google.com}

\icmlkeywords{Machine Learning, ICML,kernel approximation,Performers, Transformers, scalable,quasi Monte Carlo,random features, RFFs, PRFs, Random Fourier Features, Positive Random Features}

\vskip 0.3in
]



\printAffiliationsAndNotice{*Equal senior co-leads}

\begin{abstract}
We present \textit{Simplex Random Features} (SimRFs), a new random feature (RF) mechanism for unbiased approximation of the softmax and Gaussian kernels by geometrical correlation of random projection vectors. We prove that SimRFs provide the smallest possible mean square error (MSE) on unbiased estimates of these kernels among the class of weight-independent geometrically-coupled positive random feature (PRF) mechanisms, substantially outperforming the previously most accurate \textit{Orthogonal Random Features} (ORFs, \citealp{yu2016orthogonal}) at no observable extra cost. We present a more computationally expensive \emph{SimRFs+} variant, which we prove is asymptotically optimal in the broader family of weight-dependent geometrical coupling schemes (which permit correlations between random vector directions and norms). In extensive empirical studies, we show consistent gains provided by SimRFs in settings including pointwise kernel estimation, nonparametric classification and scalable Transformers \cite{choromanski2020rethinking}.\footnote{Code is available at \href{https://github.com/isaac-reid/simplex_random_features}{https://github.com/isaac-reid/\newline simplex\_random\_features}.}
\end{abstract}
\vspace{-6mm}
\section{Introduction}
\label{submission}

Embedding methods, which project feature vectors into a new space, are ubiquitous in machine learning. The canonical example is the Johnson-Lindenstrauss Transform (JLT) \citep{johnson1984extensions, dasgupta_jlt, sparseJLT, karnick-rf}, where a collection of high-dimensional points is embedded in a much lower dimensional space whilst (approximately) preserving their metric relationships, e.g. distances and dot-products. Another application is found in kernel approximation \citep{RF-survey, qmc-rf, pennington-rf, ionescu-rf}, where the nonlinear similarity measure (kernel) in the original space is translated to a linear kernel in the latent space. For example, a kernel $K(\cdot,\cdot): \mathbb{R}^d \times \mathbb{R}^d \rightarrow \mathbb{R} $ can be approximated using so-called \textit{random features} (RFs): randomised nonlinear transformations $\phi(\cdot): \mathbb{R}^d \rightarrow \mathbb{R}^{d'}$ constructed such that
\begin{equation}
K(\boldsymbol{x},\boldsymbol{y}) = \mathbb{E}[\widehat{K}(\boldsymbol{x,y})], \textrm{ where } \widehat{K}(\boldsymbol{x,y})\overset{\mathrm{def}}{=} \phi({\boldsymbol{x}})^{\top}\phi(\boldsymbol{y}).
\end{equation}
Provided $K$ is stationary, meaning $K(\boldsymbol{x},\boldsymbol{y}) = K(\boldsymbol{x}-\boldsymbol{y})$, we can use Bochner's theorem to write 
\begin{equation} \label{rffdef}
    K(\boldsymbol{x}-\boldsymbol{y})= \int _{\mathbb{R}^d} p(\boldsymbol{w}) e^{i \boldsymbol{w}^{\top}(\boldsymbol{x}-\boldsymbol{y})} \text{d}^d\boldsymbol{w},
\end{equation}
where $p(\textbf{w})$ is the Fourier transform of $K$. If $K$ is positive semidefinite, $p(\textbf{w})$ is non-negative so we can treat it as a probability density. This invites Monte Carlo (MC) sampling, yielding \textit{Random Fourier Features} (RFFs) of the following form, where vectors $\boldsymbol{w}_i$ are sampled from $p(\boldsymbol{w})$, $m$ is their number and $\odot$ denotes concatenation \citep{rahimi2007random, rksink}:
\begin{equation} \label{rffdef2}
\begin{multlined}
    \phi_{\mathrm{RFF}}(\boldsymbol{z}) \overset{\mathrm{def}}{=} \sqrt{\frac{1}{m}} (\odot_{i=1}^{m}[\sin(\boldsymbol{w}_i^{\top} \boldsymbol{z}),\cos(\boldsymbol{w}_i^{\top} \boldsymbol{z}) ])^{\top}.
    \end{multlined}
\end{equation}

Furthermore, if $K$ is a Gaussian kernel, defined by
\begin{equation}
K_{\mathrm{gauss}}(\boldsymbol{x},\boldsymbol{y})\overset{\mathrm{def}}{=}\exp(-\frac{\|\boldsymbol{x}-\boldsymbol{y}\|_2^{2}}{2}),
\end{equation}
random vectors $\boldsymbol{w}_{i}$ are sampled from the multivariate Gaussian distribution $\mathcal{N}(0,\mathbf{I}_{d})$. Another kernel, of key interest in Transformer architectures \citep{vaswani-tr, choromanski2020rethinking}, is the so-called \textit{softmax kernel}: 
\begin{equation}
K_{\mathrm{smax}}(\boldsymbol{x},\boldsymbol{y}) \overset{\mathrm{def}}{=} \exp(\boldsymbol{x}^{\top}\boldsymbol{y}).
\end{equation}
Since $K_{\mathrm{gauss}}(\boldsymbol{x},\boldsymbol{y}) = K_{\mathrm{smax}}(\boldsymbol{x},\boldsymbol{y}) \exp(-\frac{x^2}{2} - \frac{y^2}{2})$, RF mechanisms for the Gaussian kernel can be readily converted into the corresponding mechanism for softmax and vice versa \cite{likhosherstov2022chefs}. Our results will hence apply to both settings. For brevity, we will mostly refer to $K_\mathrm{gauss}$. 

However, as noted in \cite{choromanski2020rethinking}, RFFs lead to unstable training of implicit linear-attention Transformers. The authors address this by proposing \textit{Positive Random Features} (PRFs), defined by
\begin{equation} \label{eq:def_gauss_kern}
K_{\mathrm{gauss}}(\boldsymbol{x},\boldsymbol{y}) = 
\mathbb{E}[\phi_{\mathrm{PRF}}(\boldsymbol{x})^{\top}\phi_{\mathrm{PRF}}(\boldsymbol{y})],
\end{equation}
where for $\boldsymbol{w}_{1},...,\boldsymbol{w}_{m} \sim \mathcal{N}(0,\mathbf{I}_{d})$, 
\begin{equation}\label{eq:def_prf}
\phi_{\mathrm{PRF}}(\boldsymbol{z})\overset{\mathrm{def}}{=}
\sqrt{\frac{1}{m}}\exp(-\|\boldsymbol{z}\|_2^{2})(\odot_{i=1}^{m}[\exp(\boldsymbol{w}_{i}^{\top}\boldsymbol{z})])^{\top}.
\end{equation}

The straightforward implementation of PRFs (and RFFs) draws $\boldsymbol{w}_{i}$ independently -- a strategy we refer to as \emph{IIDRFs}. However, the isotropy of the Gaussian distribution permits us to entangle different $\boldsymbol{w}_{i}$ to be exactly orthogonal\footnote{All $\boldsymbol{w}_i$ can be orthogonal if $m \leq d$. If $m>d$ we construct ensembles of independent orthogonal blocks.} whilst preserving the Gaussian marginal distributions $\boldsymbol{w}_{i} \sim \mathcal{N}(0, 
\mathbf{I}_{d})$ \cite{yu2016orthogonal}. This mechanism is referred to as \textit{Orthogonal Random Features} (ORFs), and is an example of a \emph{weight-independent geometrically-coupled RF mechanism}.
\begin{definition}
Consider the random  vectors $\{\boldsymbol{w}_i | i \leq m\} \subset \mathbb{R}^d$, which can be described by norms $w_i =\|\boldsymbol{w}_i \|_2$ and directions $\widehat{\boldsymbol{w}}_i = \frac{\boldsymbol{w}_i}{\|\boldsymbol{w}_i \|_2}$. An RF mechanism is described as \emph{geometrically-coupled} if the norms of random vectors $\{w_i\} $ are independent, but the directions $\{\widehat{\boldsymbol{w}}_i\}$ are permitted to be correlated with one another and with the norms $\{w_i\}$. Such a coupling is \emph{weight-independent} under the further restriction that directions $\{ \widehat{\boldsymbol{w}}_i\}$ are independent of the norms $\{w_i\}$.
\end{definition}
Unless otherwise stated, all coupling mechanisms considered in this work will be geometrical. ORFs provide a lower mean squared error (MSE) on Gaussian kernel approximation than IIDRFs \cite{yu2016orthogonal, choromanski2020rethinking}, though for RFFs only at asymptotically large $d$. ORFs are used in a broad range of applications including kernel ridge regression and Transformers. In the latter case,  they offer linear (cf. quadratic) space- and time-complexity of the attention module, enabling efficient long-range attention modelling as part of the so-called Performer architecture \cite{choromanski2020rethinking}. Sec. \ref{sec:related_work} details further applications beyond Gaussian and softmax kernel estimation. Recently \citet{likhosherstov2022chefs} showed that further MSE reduction (for fixed $m$ and preserving unbiasedness) can be achieved by collecting light data statistics. RFs can also be applied with more computationally expensive pre-processing to improve accuracy in downstream tasks \cite{trokicic}, but they no longer approximate the Gaussian kernel. 

However, the following question remains open:
\textit{do ORFs provide the lowest possible MSE on unbiased estimates of the Gaussian kernel among the class of weight-independent geometrically-coupled PRF mechanisms?}

Here, we comprehensively answer this question, finding that ORFs are \emph{not} optimal. We derive the optimal mechanism, coined \textit{Simplex Random Features} (SimRFs), and show that it substantially outperforms ORFs at close to no extra computational cost. We also consider the broader family of weight-\emph{dependent} geometrically-coupled PRFs, where random vector directions $\{\widehat{\boldsymbol{w}}_i\}$ can be correlated with norms $\{w_i\}$, and present a \emph{SimRFs+} variant which we prove is asymptotically optimal in this more general class. Our empirical studies demonstrate the consistent gains provided by SimRFs in diverse settings, including pointwise kernel estimation, nonparametric classification and scalable Transformers \cite{choromanski2020rethinking}.

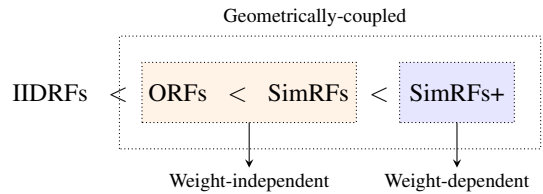
\begin{figure}
    \centering
    \begin{tikzpicture}[baseline=(current  bounding  box.center), scale=0.5]

\def\eps{1.5}
\def\yeps{2.5}
\def\boxp{0.35}

\draw [densely dotted] (-3.7,-1.5) rectangle (7.5,1.5);

\draw [densely dotted,fill=orange!10] (-3.1,-0.8) rectangle (2.6,0.8);
\draw [-stealth](-0.25,-0.8) -- (-0.25,-2);

\draw [densely dotted,fill=blue!10] (3.75,-0.8) rectangle (6.8,0.8);
\draw [-stealth](5.275,-0.8) -- (5.275,-2);
\node[scale=0.95] at (0,0) {IIDRFs $\hspace{2mm}<\hspace{2mm} $    ORFs  $\hspace{2mm}<\hspace{2mm}$    SimRFs    $\hspace{2mm}<\hspace{2mm}$    SimRFs+};
\node[scale=0.75] at (1.5,2) {Geometrically-coupled};
\node[scale=0.75] at (-0.25,-2.4) {Weight-independent};
\node[scale=0.75] at (5.275,-2.4) {Weight-dependent};

\end{tikzpicture}
    \caption{Schematic of performance of RF mechanisms described in this manuscript. SimRFs and SimRFs+ are novel.}
    \label{fig:comparison_schematic}
\end{figure}

In more detail, our principal contributions are as follows:
\begin{enumerate}
\item In Sec. \ref{sec:algorithm}, we introduce SimRFs and prove that they provide the lowest kernel estimator MSE of any weight-independent geometrically-coupled PRF mechanism, outperforming the previously most accurate ORFs. We demonstrate that a fast, simple scheme applying minor alterations to SimRFs yields SimRFs+: a marginally better weight-dependent mechanism. See Fig. \ref{fig:comparison_schematic}. 
\item In Sec. \ref{sec:theory}, we provide novel theoretical results to add insight to the discussion in Sec. \ref{sec:algorithm}. They may be of independent interest. 
We derive the first non-asymptotic \textbf{closed-form} formulae for the MSE for PRFs in the IIDRF, ORF and SimRF settings, and show how it is straightforward to generalise some of these forms to RFFs. This allows us to precisely quantify how much the kernel estimator MSE can be suppressed by geometrical coupling. 
We also compare the time- and space-complexities of the different PRF mechanisms and describe a faster, approximate implementation.

\item In Sec. \ref{experiments}, we support our theoretical results with comprehensive experiments, demonstrating the superiority of SimRFs over ORFs and IIDRFs. We empirically confirm that they  offer lower kernel estimator MSE, and find that this translates to better downstream performance in nonparametric classification tasks (Sec. \ref{nonparam_class}) and scalable Transformers (Sec. \ref{exp_performers}).
\end{enumerate}
\vspace{-3mm}
Proofs not provided in the main body are in Appendix \ref{app:all_proofs_appendix}.

\vspace{-3mm}
\section{Related Work}
\label{sec:related_work}

The literature on \textit{structured} RFs, where random vectors are conditionally dependent, is extensive \cite{chazelle, liberty-1, liberty-2, fastfood, binary-embeds}. ORFs were first proposed for nonlinear kernel estimation in \cite{yu2016orthogonal}, where the authors derived strict asymptotic gains from ORFs compared to IIDRFs when using RFFs for Gaussian kernel approximation. We refer to this phenomenon -- the supression of kernel estimator MSE when random features are conditioned to be orthogonal -- as \textit{the orthogonality gap}.

Further progress towards an understanding of the orthogonality gap was provided in \cite{grfs}, where the authors introduced and studied the so-called \textit{charm property} of stationary kernels. However, a rigorous mathematical analysis in the non-asymptotic setting remained out of reach. In \cite{choromanski2017unreasonable}, the authors showed the superiority of ORFs over IIDRFs for angular kernel estimation in any $d$ (not just asymptotic) and conducted an extensive analysis of the linear (dot-product) kernel, but they did not address stationary kernels. The authors of \cite{dpp-orfs} used the lens of \textit{determinantal point processes} and the negative dependence property \cite{dpps} to explore the efficacy of ORFs.

ORFs are used with PRFs in Performers \cite{choromanski2020rethinking, schlag, shandali, slimperformer, dubey_kernel, performer-mpc}: a recently-proposed class of efficient Transformer \citep{reformer, routing_transformer} that can be applied to ultra-long sequences or to expedite inference on regular-size sequences.
\vspace{-3mm}

\section{Simplex Random Features (SimRFs)}
\label{sec:algorithm}

In this section, we describe our core contributions. 

We begin by presenting Simplex Random Features (SimRFs). In analogy to the square orthogonal block, we define the so-called \textit{simplex block}, consisting of $d$ $d$-dimensional random vectors $\{\boldsymbol{w}_{i} | i \leq d\}$. In practical applications where $m>d$ random features are needed, multiple simplex blocks are constructed independently. 
\begin{figure}
\centering
\begin{tikzpicture}[baseline=(current  bounding  box.center), scale=0.5]

\def\eps{3.5}
\def\yeps{2.5}
\def\boxp{0.35}

\draw[thick, fill=black] (0,\yeps) circle (0.1);
\draw[thick] (0,\yeps)--(0,1+\yeps);
\path (0,1+\yeps) pic[rotate=180,scale=0.1] {arrowhead};
\draw[thick] (0,\yeps)--(1.4*0.25,1.3+\yeps);
\path (1.4*0.25,1.3+\yeps) pic[rotate=165,scale=0.1] {arrowhead};

\draw[thick, fill=black] (0,0) circle (0.1);
\draw[thick] (0,0)--(0,1*1.3);
\path (0,1*1.3) pic[rotate=180,scale=0.1] {arrowhead};
\draw[ultra thick] (0,0)--(0.259*0.8,-0.966*0.8);
\path (0.259*0.8,-0.966*0.8) pic[rotate=15,scale=0.1] {arrowhead};
\draw[thick][densely dotted] (0,0)--(1.5,-0.3);
\path (1.5,-0.3) pic[rotate=78.7,scale=0.1] {arrowhead};

\draw[thick, fill=black] (\eps,\yeps) circle (0.1);
\draw[thick] (\eps,\yeps)--(\eps,1+\yeps);
\path (\eps,1+\yeps) pic[rotate=180,scale=0.1] {arrowhead};
\draw[thick] (\eps,\yeps)--(1.4+\eps,\yeps);
\path (1.4+\eps,\yeps) pic[rotate=90,scale=0.1] {arrowhead};
\draw (\eps,\yeps+\boxp)--(\boxp+\eps,\yeps+\boxp) -- (\boxp+\eps,\yeps);

\draw[thick] (\eps,0)--(\eps,1*1.3);
\path (\eps,1*1.3) pic[rotate=180,scale=0.1] {arrowhead};
\draw[thick](\eps,0)--(\eps+1.5,0);
\path (\eps+1.5,0) pic[rotate=90,scale=0.1] {arrowhead};
\draw[thick, fill=white] (\eps,0) circle (0.2);
\draw[thick, fill=black] (\eps,0) circle (0.05);
\draw (\eps,\boxp)--(\eps+\boxp,\boxp) -- (\eps+\boxp,0);

\draw[thick, fill=black] (2*\eps,\yeps) circle (0.1);
\draw[thick] (2*\eps,\yeps)--(2*\eps,\yeps+1.3);
\path (2*\eps,\yeps+1.3) pic[rotate=180,scale=0.1] {arrowhead};
\draw[thick] (2*\eps,\yeps)--(2*\eps,\yeps-0.8);
\path (2*\eps,\yeps-0.8) pic[rotate=0,scale=0.1] {arrowhead};

\draw[thick, fill=black] (2*\eps,0) circle (0.1);
\draw[thick] (2*\eps,0)--(2*\eps,1.2);
\path (2*\eps,1.2) pic[rotate=180,scale=0.1] {arrowhead};
\draw[thick] (2*\eps,0)--(2*\eps+0.779,-0.45);
\path (2*\eps+0.779,-0.45) pic[rotate=60,scale=0.1] {arrowhead};
\draw[thick] (2*\eps,0)--(2*\eps-1.386,-0.8);
\path (2*\eps-1.386,-0.8) pic[rotate=-60,scale=0.1] {arrowhead};
 \draw (2*\eps,\boxp) arc (90:-30:\boxp);
\node[scale=0.75] at (2*\eps+1,0.6) {$120^\circ$};

\node[scale=1] at (0,2*\yeps) {IIDRFs};
\node[scale=1] at (\eps,2*\yeps) {ORFs};
\node[scale=1] at (2*\eps,2*\yeps) {SimRFs};
\node[scale=1] at (-\eps,\yeps) {$d=2$};
\node[scale=1] at (-\eps,0) {$d=3$};
\end{tikzpicture}
\caption{\small{Schematic of different geometrical couplings for small $d$. Dotted lines have a component into the plane of the paper, thick lines have a component out, and $\odot$ is purely out (i.e. perpendicular to the paper's plane). With IIDRFs, the respective orientations of vectors are chosen independently. With ORFs, we condition the vectors to be perpendicular. With SimRFs, they subtend angles $\theta=\arccos(-\frac{1}{d-1})$. Intuitively, conditioning the vectors to subtend fixed, obtuse angles means they `explore' $\mathbb{R}^d$ better, suppressing the kernel estimator MSE. All norms are drawn independently from a $\chi_d$-distribution.}}
\vspace{-4.5mm}
\label{fig:scheme}
\end{figure}
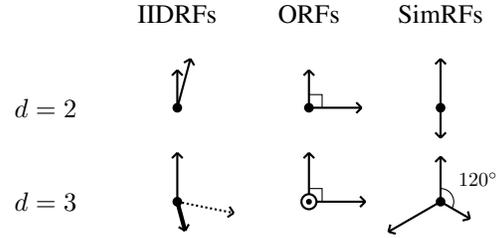

Instead of being orthogonal, the rows of the simplex block point towards the vertices of a $d-1$-dimensional simplex embedded in $d$-dimensional space, subtending angles $\theta = \arccos(-\frac{1}{d-1})$. The entire simplex (or, equivalently, the vector it operates on) is randomly rotated to preserve isotropy, and the rows are independently renormalised by weights $w_i \sim \chi_d$ such that they are marginally Gaussian. Explicitly, we define the simplex block $\mathbf{W}_\textrm{simp} \in \mathbb{R}^{d \times d}$ by
\begin{equation} \label{simplex_block_eq}
\mathbf{W}_\textrm{simp} = \mathbf{DSR}
\end{equation}
where $\mathbf{D} \in \mathbb{R}^{d \times d} = \mathrm{diag}(w_i)$ with $w_i$ sampled from a $\chi_d$-distribution. $\mathbf{R} \in \mathbb{R}^{d \times d}$ is a random orthogonal matrix drawn from Haar measure on $\mathrm{O}(d)$, the group of orthogonal matrices in $\mathbb{R}^{d\times d}$, constructed e.g by Gram-Schmidt orthogonalisation of an unstructured Gaussian matrix \cite{yu2016orthogonal}. The rows $\boldsymbol{s}_i$ of the simplex projection matrix $\mathbf{S} \in \mathbb{R}^{d \times d}$ are given by the unit vectors 
\begin{equation}
\label{eq:simplex_vectors}
\boldsymbol{s}_i = \begin{cases}
     \sqrt{\frac{d}{d-1}} \textbf{e}_i - \frac{\sqrt{d}+1}{(d-1)^{3/2}} (1,...,1,0)^{\top}  & \text{for}\ 1 \leq i < d \\
      \frac{1}{\sqrt{d-1}}(1,1,...,1,0)^{\top} & \text{for}\ i = d \\
    \end{cases}
\end{equation}
which are manifestly normalised and subtend obtuse angles. Fig. \ref{fig:scheme} visualises the  different geometrical couplings of IIDRFs, ORFs and SimRFs in low data dimensionality $d$.

\subsection{RF-Conformity and SimRFs vs ORFs}
Recalling again that the Gaussian and softmax kernels are readily interchanged, we focus on $K_\mathrm{gauss}$ without loss of generality. We begin by defining the \emph{RF-conformity}.
\begin{definition}
The RF-conformity, $\rho(\boldsymbol{x},\boldsymbol{y})$, is given by \small
\begin{equation} \label{eq:def_rf_conformity}
 \rho(\boldsymbol{x}, \boldsymbol{y})\overset{\mathrm{def}}{=}
\frac{\Gamma(\frac{d}{2})}{m(m-1)} \sum_{i,j \neq i} \mathbb{E}_{w_{ij}}\left( \sum_{k=0}^\infty \frac{v^{2k} w_{ij} ^{2k}}{2^{2k} k! \Gamma(k+\frac{d}{2})} \right),
\end{equation} \normalsize
with $w_{ij}=\|\boldsymbol{w}_{i}+\boldsymbol{w}_{j}\|_{2}$, $v=\|\boldsymbol{x}+\boldsymbol{y}\|_{2}$ for $\boldsymbol{x},\boldsymbol{y}\in\mathbb{R}^d$, $\Gamma$ the Gamma-function and $m$ the no. random vectors $\boldsymbol{w}_i$.
\end{definition}
$\rho(\boldsymbol{x},\boldsymbol{y})$ depends on correlations induced between random vector directions. It is bigger when random vectors point in similar directions, `exploring' $\mathbb{R}^d$ less effectively. In Appendix \ref{app:conformity}, we prove the following important result.
\begin{theorem}[MSE depends on RF-conformity]
\label{thm:rf-conformity} 
For PRFs, the $\mathrm{MSE}$ of the unbiased estimator $\widehat{K}(\boldsymbol{x},\boldsymbol{y})$ is given by
\small
\begin{equation} \label{mse_and_conformity}
\begin{multlined}
\text{\emph{MSE}}(\widehat{K})= \frac{e^{-2x^2-2y^2}}{m}\left ((e^{2v^2}- e^{v^2}) \right.\\\left.
+ (m-1)(\rho(\boldsymbol{x},\boldsymbol{y}) - e^{v^2}) \right).
\end{multlined}
\end{equation} 
\normalsize
That is, the MSE is an increasing function of the RF-conformity.
\end{theorem}
 For any $w_i,w_j$, SimRFs give strictly smaller values of $w_{ij}$ than ORFs because the random vectors subtend a bigger angle. Explicitly, $w_{ij} = (w_i^2 + w_j^2 + 2w_iw_j\cos\theta)^{1/2}$ is smaller when $\cos \theta = -\frac{1}{d-1}$ (SimRFs) compared to when $\cos \theta=0$ (ORFs). This leads to smaller values of $\rho(\boldsymbol{x},\boldsymbol{y})$, which immediately implies the following important result.
\begin{corollary}[SimRFs outperform ORFs]
For PRFs, the kernel estimator MSE obtained with SimRFs is strictly lower than with ORFs for arbitrary data dimensionality $d$.
\end{corollary}
In fact, we are able to make the following substantially stronger statement, proved in Appendix \ref{app:equal_weights}.
\begin{theorem}[SimRFs optimal for weight-independent geometrical coupling] \label{thm:simplex_optimal}
Supposing that $d$ random vector norms $\{w_i | i\leq d\}$ are i.i.d., SimRFs constitute the \textbf{best possible weight-independent geometrical coupling mechanism}, giving the lowest possible PRF kernel estimator MSE.
\end{theorem}

\subsection{SimRFs+} \label{simrfs+_sec}
Now we consider the broader family of weight-\emph{dependent} geometrical coupling mechanisms, where random vector directions $\{\hat{\boldsymbol{w}}_i\}$ are permitted to be correlated with norms $\{w_i\}$. In particular, given $d$ vectors $\{\boldsymbol{w}_i\}$ of known norms (from $d$ draws of $\chi_d$), we would like to arrange them in $d$-dimensional space in order to minimise the sum\footnote{We remove the expectation value because, given a fixed set of norms, assigning any probability mass to suboptimal configurations will increase the RF-conformity in expectation -- that is, the best geometrical coupling between vectors of known magnitudes $\{w_i\}$ is deterministic.} 
\begin{equation} \label{eq:simp+_obj}
 \rho(\boldsymbol{x}, \boldsymbol{y})=
\frac{\Gamma(\frac{d}{2})}{m(m-1)} \sum_{i,j \neq i} \left( \sum_{k=0}^\infty \frac{v^{2k} w_{ij} ^{2k}}{2^{2k} k! \Gamma(k+\frac{d}{2})} \right).
\end{equation} \normalsize
One brute-force approach is to parameterise each of the $d$ random vector directions in hyperspherical coordinates and use an off-the-shelf numerical optimiser (e.g. $\mathrm{scipy.optimize}$). This is prohibitively slow, and moreover the solution has data-dependence via $v=\|\boldsymbol{x}+\boldsymbol{y}\|_2$ which frustrates the method's scalability: the optimisation needs to be carried out pairwise for every $(\boldsymbol{x},\boldsymbol{y})$, which undermines our ability to quickly evaluate $\widehat{K}(\boldsymbol{x},\boldsymbol{y}) = \phi(\boldsymbol{x})^{\top} \phi(\boldsymbol{y})$ for any given pair of input vectors. However, the numerical approach does benchmark the lowest possible RF-conformity that can be achieved with weight-dependent geometrical coupling.

The generic \emph{analytic} minimisation of Eq. \ref{eq:simp+_obj} is challenging, and solutions will suffer the same $v$-dependence described above, so we instead consider a tractable approximation. Dropping constant prefactors for clarity, the first few terms from Eq. \ref{eq:def_rf_conformity} are given by:
\begin{equation} \label{ensemble_expand_out}
\begin{aligned}
\small
 \sum_{i, j\neq i} \mathbb{E}_{w_{ij}}\left( \frac{1}{\Gamma(\frac{d}{2})} + \frac{v^2 w_{ij}^2 }{4 \Gamma(\frac{d}{2}+1)} + \frac{v^4 w_{ij}^4 }{32 \Gamma(\frac{d}{2}+2)}+... \right)
 &\\ = \frac{1}{\Gamma(\frac{d}{2})}\sum_{i, j\neq i} 1 + \tau \left(1 + \frac{v^2}{8} \frac{\Gamma(\frac{d}{2} +1)}{\Gamma(\frac{d}{2}+2)} \frac{\mathbb{E}(w_{ij}^4)}{\mathbb{E}(w_{ij}^2)}  + ... \right) 
 \normalsize
\end{aligned}
\end{equation}
with $\tau=\frac{\Gamma(\frac{d}{2})v^2 \mathbb{E}(w_{ij}^2) }{4 \Gamma(\frac{d}{2}+1)}$. The precise value of $\frac{\mathbb{E}(w_{ij}^4)}{\mathbb{E}(w_{ij}^2)}$ will depend on the geometrical coupling scheme employed, but for the types we have considered we generally expect it to scale as $\sim d$, with some constant prefactor\footnote{For example, with orthogonal coupling $\frac{\mathbb{E}(w_{ij}^4)}{\mathbb{E}(w_{ij}^2)} = \frac{\mathbb{E}(w_i^4 + w_j^4 + 2w_i^2w_j^2)}{\mathbb{E}(w_i^2 +w_j^2)} = \frac{\mathbb{E}(w_i^4)}{\mathbb{E}(w_i^2)} + \mathbb{E}(w_i^2) = 2\frac{\Gamma(\frac{d}{2} + 2)}{\Gamma(\frac{d}{2}+1)} + 2\frac{\Gamma(\frac{d}{2} + 1)}{\Gamma(\frac{d}{2})} \sim d$, where we took moments of the $\chi_d$ distribution. We can perform similar analyses in the i.i.d. and simplex cases. }. Therefore the sum in Eq. \ref{eq:def_rf_conformity} can be approximated by: 
\begin{equation}
 \frac{1}{\Gamma(\frac{d}{2})}\sum_{i,j\neq i} 1 + \frac{\Gamma(\frac{d}{2})v^2 \mathbb{E}(w_{ij}^2) }{4 \Gamma(\frac{d}{2}+1)}\left(1 + \mathcal{O}(v^2) + ... \right)  .
\end{equation}
In the limit of small $v$, this invites us to truncate the sum at $k=1$, dropping the $\mathcal{O}(v^2)$ terms. Omitting additive constants, we are left with the approximate objective
\begin{equation}\label{eq:truncated_obj}
\tilde{\rho}(\boldsymbol{x},\boldsymbol{y})=\frac{\Gamma(d/2) v^2}{4m(m-1)\Gamma(1+d/2)} \sum_{i,j \neq i} w_{ij}^2,
\end{equation}
the physical analogue of which is the Heisenberg Hamiltonian with different coupling constants between different spin pairs. This is exactly minimised by
\begin{equation} \label{minimised_conditions_2_main}
 \boldsymbol{w}_i=-\frac{\sum_{j\neq i}  \boldsymbol{w}_j}{\|\sum_{j\neq i}  \boldsymbol{w}_j\|_2} w_i  \hspace{5ex} i=1,...,d
\end{equation}
where each random vector points away from the resultant of all the others (see Appendix \ref{app:simp_plus_min} for details). Fig. \ref{fig:splusfig-main} captures this essential difference between SimRFs and SimRFs+: in the latter case, vectors with larger norms subtend bigger angles. Empirically, we find that the iterative update scheme
\begin{equation} \label{simpler_updates_main}
\boldsymbol{w}_i \leftarrow -\frac{\sum_{j\neq i} \boldsymbol{w}_j}{\|\sum_{j\neq i} \boldsymbol{w}_j\|_2} w_i
\end{equation}
converges to Eq. \ref{minimised_conditions_2_main} quickly (after a small number of passes through the set of $d$ vectors), especially if we initialise in the near-optimal simplex geometry.
Conveniently, the solution has no $v$-dependence and is therefore scalable: the optimisation needs to be carried out for every draw of weights $\{w_i$\} but \emph{not} every pair of data points $(\boldsymbol{x},\boldsymbol{y})$. We refer to this mechanism of weight-dependent geometrical coupling as \emph{SimRFs+}, and emphasise that it is asymptotically optimal (in the sense of minimising $\rho(\boldsymbol{x},\boldsymbol{y})$) in the $v\ll1$ limit. 
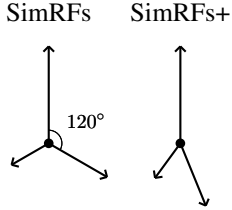
\begin{figure}
\centering
\begin{tikzpicture}[baseline=(current  bounding  box.center), scale=0.5]
\def\eps{3.5}
\def\yeps{3.5}
\def\boxp{0.35}
\draw[thick, fill=black] (0,0) circle (0.1);
\draw[thick] (0,0)--(0,2.58);
\path[thick] (0,2.58) pic[rotate=180,scale=0.1] {arrowhead};
\draw[thick] (0,0)--({1.79*sin(120)},{1.79*cos(120)});
\path[thick] ({1.79*sin(120)},{1.79*cos(120)}) pic[rotate=60,scale=0.1] {arrowhead};
\draw[thick] (0,0)--({1.14*sin(240)},{1.14*cos(240)});
\path ({1.14*sin(240)},{1.14*cos(240)}) pic[rotate=-60,scale=0.1] {arrowhead};
 \draw (0,\boxp) arc (90:-30:\boxp);
\node[scale=0.75] at (0+1,0.6) {$120^\circ$};
\draw[thick, fill=black] (\eps,0) circle (0.1);
\draw[thick] (\eps,0)--(\eps,2.58);
\path[thick] (\eps,2.58) pic[rotate=180,scale=0.1] {arrowhead};
\draw[thick] (\eps,0)--({\eps+1.79*sin(157.8)},{1.79*cos(157.8)});
\path[thick] ({\eps+1.79*sin(157.8)},{1.79*cos(157.8)}) pic[rotate=60-37.8,scale=0.1] {arrowhead};
\draw[thick] (\eps,0)--({\eps+1.14*sin(216.2)},{1.14*cos(216.2)});
\path ({\eps+1.14*sin(216.2)},{1.14*cos(216.2)}) pic[rotate=-60+23.8,scale=0.1] {arrowhead};
 \draw (0,\boxp) arc (90:-30:\boxp);
\node[scale=0.75] at (0+1,0.6) {$120^\circ$};
\node[scale=1] at (0,\yeps) {SimRFs};
\node[scale=1] at (\eps,\yeps) {SimRFs+};
\end{tikzpicture}
\caption{\small{With SimRFs, random vectors are geometrically correlated such that all pairs subtend an equal angle $\theta = \arccos(-\frac{1}{d-1})$. With SimRFs+, random vectors with bigger norms subtend bigger angles, guaranteeing smaller kernel estimator MSE when $v$ is sufficiently small.}}
\label{fig:splusfig-main}
\end{figure}

\begin{figure}[H] 
\centering
\vspace{-3mm}
  \includegraphics{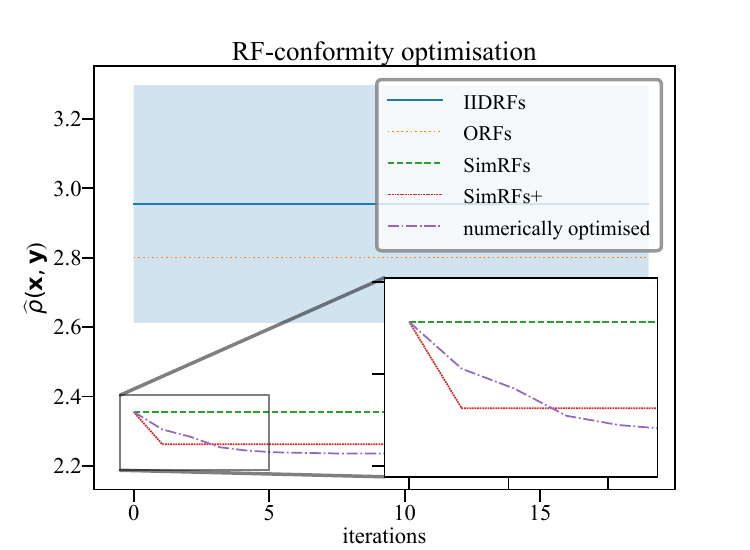}
  \caption{Comparison of the RF-conformity defined in Eq. \ref{eq:def_rf_conformity} (lower is better) for a \emph{single random draw} of norms $\{ w_i \}$, $v = \|\boldsymbol{x} + \boldsymbol{y}\|_2 = 1$ and $d=6$. IIDRFs, ORFs, SimRFs and SimRFs+ are implemented as described in the main text. `Numerically optimised' uses an off-the-shelf numerical optimiser to arrange vectors to minimise the RF-conformity: a scheme which is too computationally inefficient to be practical but benchmarks the lowest possible value. Any improvements above SimRFs using weight-dependent geometrical coupling are marginal. The IIDRF value is averaged over 100 random couplings of fixed weights, and the shaded region gives 1 standard deviation.} 
  \label{corr_inset_main}
\end{figure}

 Fig. \ref{corr_inset_main} compares the RF-conformity of the mechanisms we have considered, as well as the outcome of the inefficient numerical optimisation. 
 The additional benefits of weight-dependent coupling are marginal: SimRFs+ access only slightly lower conformity than SimRFs at the expense of an extra optimisation step of time-complexity $\mathcal{O}(d^3)$. This gives context to the excellent performance of SimRFs; they can compete with members of a much broader class at a fraction of the computational cost. We also note that the minimisation of the truncated objective (SimRFs+) is a good approximation to the minimisation of the true objective (`numerically optimised'), accessing comparably small values of $\rho$. Informally, SimRFs+ are close to optimal among the class of weight-dependent geometrically-coupled PRF mechanisms.

\vspace{-5mm}
\section{From ORFs to SimRFs: the Theory}
\label{sec:theory}
This section provides more detailed theoretical analysis to add insight to the results of Sec. \ref{sec:algorithm}. It can safely be omitted on a quick reading. We derive analytic expressions for the RF-conformity $\rho(\boldsymbol{x},\boldsymbol{y})$, and therefore the kernel estimator MSE, for IIDRFs, ORFs and SimRFs. This allows us to quantitatively compare the performance of different coupling mechanisms. As before, we specialise to $K_\mathrm{gauss}$. Detailed proofs are provided in Appendix \ref{app:all_proofs_appendix}. 

We have seen that RF-conformity depends on an expectation value over $w_{ij}=\|\boldsymbol{w}_i + \boldsymbol{w}_j \|_2$. This motivates us to begin with the following auxiliary lemma.
\begin{lemma}[IIDRF conformity]
\label{iidrf_conformity}
When random vectors $\boldsymbol{w}_i,\boldsymbol{w}_j \in \mathbb{R}^d$ are i.i.d. (IIDRFs), the probability distribution $p(w_{ij})$ with $w_{ij}=\|\boldsymbol{w}_i + \boldsymbol{w}_j\|_2$ is given by 
\begin{equation} \label{eq:iid_pdf}
     p_\textrm{i.i.d.}(w_{ij}) = \frac{w_{ij}^{d-1} e^{-w_{ij}^2/4}}{2^{d-1} \Gamma(\frac{d}{2})} 
\end{equation}
which induces an RF-conformity 
\begin{equation} \label{eq:iidrf_conformity}
\rho_{\mathrm{IIDRF}}(\boldsymbol{x},\boldsymbol{y}) = e^{v^2}
\end{equation}
where $\boldsymbol{x},\boldsymbol{y}\in\mathbb{R}^d$ and $v=\|\boldsymbol{x}+\boldsymbol{y}\|_2$.
\end{lemma}
Now we make the following important observation. 
\begin{lemma}[PDF for vectors subtending $\theta$]\label{angle_pdf}
Supposing random vectors $\boldsymbol{w}_{i},\boldsymbol{w}_j$ are marginally Gaussian but are conditioned to subtend a fixed angle $\theta$, the probability distribution $p_\theta(w_{ij})$, is given by 
\begin{equation}
   \frac{w^{2d-1}}{2^{d-2} \Gamma(\frac{d}{2})^2} \int_{\phi =0}^{\pi/2} \text{d}\phi ( \sin \phi \cos \phi)^{d-1} \frac{e^{-\frac{w^2}{2(1 + \sin 2 \phi \cos \theta)}}}{(1 + \sin 2 \phi \cos \theta)^{d}}. 
    \label{eq:theta_proby_dist}
\end{equation}
\end{lemma}
ORFs and SimRFs correspond to special instances of this with $\cos \theta = 0$ and $\cos \theta = -\frac{1}{d-1}$, respectively. It is instructive to observe that, in the orthogonal case, the distribution reduces to the $\chi_{2d}$-distribution. The probability distribution $p_\theta(w_{ij})$ induces an RF-conformity
\begin{equation}
\begin{multlined}
    \rho_\theta(\boldsymbol{x},\boldsymbol{y}) = \frac{1}{2^{d-1}\Gamma(\frac{d}{2})} \int_0^{\pi}\text{d}\phi ( \sin \phi )^{d-1} \\ \cdot \sum_{k=0}^\infty \frac{v^{2k}(1 + \sin\phi\cos\theta)^k}{2^{k} k! \Gamma(k+\frac{d}{2})} \Gamma(k+d).
    \label{eq:theta_conformity}
\end{multlined}
\end{equation}
Inspecting the form closely, we see that every term in the sum over $k$ is proportional to the integral
\begin{equation} \label{binexp_main}
    \int_0^{\pi}\text{d}\phi ( \sin \phi )^{d-1} (1 + \sin\phi\cos\theta)^k
\end{equation}
which is strictly smaller for $\cos{\theta}<0$ compared to $\cos{\theta}=0$ (since $\sin\phi$ is nonnegative everywhere in the domain). Since every term in the sum is positive, we immediately conclude that for PRFs the conformity of SimRFs is strictly smaller than ORFs, and hence the MSE is smaller. We already derived this in Sec. \ref{sec:algorithm}, but are now also able to provide the following closed forms.
\begin{theorem}[ORF and SimRF conformity closed forms]\label{prf_gap_1_main} For PRFs with $\boldsymbol{x},\boldsymbol{y}\in\mathbb{R}^d$, the RF-conformity of ORFs is
\begin{equation} \label{eq:orf_conformity}
\rho_{\mathrm{ORF}}(\boldsymbol{x},\boldsymbol{y}) =\frac{\Gamma(\frac{d}{2})}{\Gamma(d)} \sum_{k=0}^\infty \frac{v^{2k}}{2^k k!} \frac{\Gamma(k+d)}{\Gamma(k+\frac{d}{2})}
\end{equation}
whereas the RF-conformity of SimRFs is
\begin{equation}
\begin{multlined} \label{eq:simrf_conformity}
\rho_\mathrm{SimRF}(\boldsymbol{x},\boldsymbol{y}) =\frac{\sqrt{\pi}}{\Gamma(\frac{d}{2}) 2^{d-1}}\sum_{k=0}^\infty \frac{\Gamma(k+d)}{\Gamma(k+\frac{d}{2})}  \frac{v^{2k}}{2^k} \\ \cdot \sum_{p=0}^k \left(-\frac{1}{d-1}\right)^p \frac{\Gamma(\frac{d+p}{2})}{\Gamma(\frac{d+p+1}{2})} \frac{1}{(k-p)! p!}.
\end{multlined}
\end{equation}
\end{theorem}
These results are novel. They permit the \textbf{first analytic characterisation of the difference in kernel estimator MSE between IIDRFs, ORFs and SimRFs}. We make one further observation.
\begin{corollary}[ORFs always outperform IIDRFs]\label{corr:orth_gap} 
In the PRF setting, the orthogonality gap (difference in kernel estimator MSE between IIDRFs and ORFs) is given by 
\begin{equation} \label{eq:prf_orth_gap}
\begin{multlined}
\Delta \mathrm{MSE}(\widehat{K}(\mathbf{x},\mathbf{y})) = e^{-2x^2-2y^2} \frac{m-1}{m} 
\\ \cdot \left( e^{v^2} - \frac{\Gamma(d/2)}{\Gamma(d)}\sum_{k=0}^\infty \frac{v^{2k}}{k!} \frac{\Gamma(k+d)}{\Gamma(k+d/2)}\right)
\end{multlined}
\end{equation}
where $\boldsymbol{x},\boldsymbol{y} \in \mathbb{R}^d$, $v=\|\boldsymbol{x}+\boldsymbol{y}\|_2$ and $m\leq d$ is the number of random vectors. This is positive everywhere. 
\end{corollary}
The sign of this orthogonality gap was first reported in \cite{choromanski2020rethinking} but without an accompanying closed form.

\begin{figure}[H]
\vspace{-0mm}
\centering
  \includegraphics{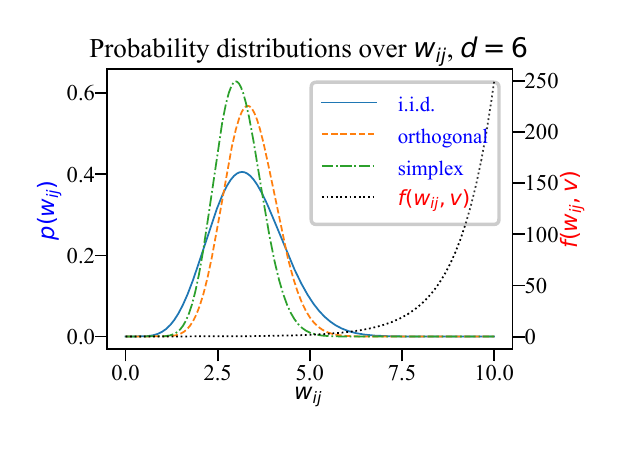}
  \vspace{-5mm}
  \caption{Probability distributions over the random variable $w_{ij}=\|\boldsymbol{w}_i+\boldsymbol{w}_j\|_2$ for IIDRFs, ORFs and SimRFs. The RF-conformity depends on the expectation of a monotonically increasing function $f(w_{ij})$. With PRFs, geometrical coupling decreases this by reducing the probability mass at large $w_{ij}$.}
\vspace{-5mm}
\label{fig:the_dists}
\end{figure}

Plotting each of derived probability distributions $p(w_{ij})$ (Eq. \ref{eq:iid_pdf} and Eq. \ref{eq:theta_proby_dist}, taking $\cos\theta=0$ and $\cos\theta=-\frac{1}{d-1}$) and noting from Eq. \ref{eq:def_rf_conformity} that the RF-conformity depends on the expectation value of the monotonically increasing function $f(w_{ij},v)=
\Gamma(\frac{d}{2})\sum_{k=0}^\infty \frac{v^{2k} w_{ij} ^{2k}}{2^{2k} k! \Gamma(k+\frac{d}{2})}$, the intuitive reason for the relative efficacy of SimRFs, ORFs and IIDRFs becomes clear: conformity is penalised by tails at large $w_{ij}$, which we suppress with geometrical coupling (Fig. \ref{fig:the_dists}).  

\subsection{Extension to RFFs}
We briefly note that, with minimal work, the preceding results for PRFs can be modified to consider RFFs. For example, the following is true.
\begin{theorem}[RFF orthogonality gap] \label{thm:rff_gap} In the RFF setting, the orthogonality gap (difference in kernel estimator MSE between IIDRFs and ORFs) is given by
\begin{equation}
\begin{multlined}
\Delta \text{\emph{MSE}}(\widehat{K}(\boldsymbol{x},\boldsymbol{y})) = \frac{m-1}{m} \left( e^{-z^2} - \right. \\ \left. \frac{\Gamma(d/2)}{\Gamma(d)} \sum_{k=0}^\infty \frac{(-z^2)^k}{2^k k!} \frac{\Gamma(k+d)}{\Gamma (k+d/2)} \right)
\end{multlined}
\end{equation}

where $\boldsymbol{x},\boldsymbol{y} \in \mathbb{R}^{d}$, $z=\|\boldsymbol{x}-\boldsymbol{y}\|_{2}$ and $m \leq d$ is the number of random vectors.
\end{theorem}
To the best of our knowledge, this result is also novel. The expression does not admit the same simple analysis as the PRF form (\ref{eq:prf_orth_gap}) because successive terms in the sum oscillate in sign, but a cursory numerical analysis reveals that the MSE of ORFs is smaller than IIDRFs up to some threshold $z_\text{crit}(d)$, the value of which diverges as $d\to\infty$. Taylor expanding our exact result in $\frac{1}{d}$ reproduces the following.

\begin{corollary}[RFF asymptotic MSE ratio, \citet{yu2016orthogonal}]\label{corr:asymptotic_rff_form} The ratio of ORF to IIDRF kernel estimator MSE is given by
\small
\begin{equation} \label{eq:asymptotic_ratio}
\frac{\text{MSE}(\widehat{K}_\text{ORF})}{\text{MSE}(\widehat{K}_\text{IIDRF})} = 1 - (m-1)\left(\frac{e^{-z^2} z^4}{d(1-e^{-z^2})^2} + \mathcal{O}\left(\frac{1}{d^2}\right)\right), 
\end{equation} \normalsize
where $\boldsymbol{x},\boldsymbol{y} \in \mathbb{R}^d$, $z=\|\boldsymbol{x}-\boldsymbol{y}\|_2$ and $m\leq d$ is the number of random features.
\end{corollary} \vspace{-2mm}
The negative subleading term shows that the RFF orthogonality gap is positive everywhere when $d\to\infty$.
 
\subsection{Implementation, Complexity and Fast SimRFs} \label{sec:fastsimrfs}
The replacement of ORFs with SimRFs is straightforward: instead of calculating random projections $\mathbf{W}\boldsymbol{x}$ using the \emph{orthogonal block} $\mathbf{W}_\textrm{ort}=\mathbf{DR}$, we use the \emph{simplex block} $\mathbf{W}_\textrm{simp}=\mathbf{DSR}$, with the matrices $\mathbf{D},\mathbf{S},\mathbf{R}\in \mathbb{R}^{d\times d}$ and the object $\boldsymbol{x}\in \mathbb{R}^d$  defined at the beginning of Sec. \ref{sec:algorithm}. By choosing the order of computation $\mathbf{D}(\mathbf{S}(\mathbf{R}\boldsymbol{x}))$, we can avoid the $\mathcal{O}(d^3)$ time complexity of computing matrix-matrix products. Both $\mathbf{D}$ and $\mathbf{S}$ support matrix-vector multiplication of time complexity $\mathcal{O}(d)$ (see Appendix \ref{app:fastsimsalg}). Generically, the time complexity to sample the random orthogonal matrix $\mathbf{R}$ is $\mathcal{O}(d^3)$ and the matrix-vector multiplication $\mathbf{R}\boldsymbol{x}$ is $\mathcal{O}(d^2)$. However, following exactly the same tricks as with ORFs, it is possible to replace $\mathbf{R}$ with a proxy $\widetilde{\mathbf{R}}$ which is \emph{approximately} sampled from the orthogonal group according to Haar measure and which supports fast matrix-vector multiplication: for example, $\mathbf{HD}$-product matrices \cite{choromanski2017unreasonable} or products of Givens random rotations \cite{butterfly}. Then the time-complexity will be limited by the computation $\widetilde{\mathbf{R}}\boldsymbol{x}$ which is subquadratic by construction (e.g. $\mathcal{O}(d\log d)$ for the examples above). We refer to this mechanism as \emph{fast SimRFs}, and show its excellent experimental performance in Appendix \ref{app:fastsimsexp}. 

SimRFs+ are implemented by $\mathbf{W}_\textrm{simp+}=\mathbf{DS'R}$, where $\mathbf{S}'$ is obtained from $\mathbf{S}$ according to the $\mathcal{O}(d^3)$ iterative optimisation scheme defined in Eq. \ref{simpler_updates_main}. This will dominate the  scaling of time-complexity if we apply fast SimRFs+. 
\vspace{-5mm}
\begin{table}[H]
\caption{Time complexities of RF-mechanisms and their fast variants.}
\label{tab:splus_table}\vspace{-0mm}
\begin{center}
\begin{small}
\begin{tabular}{l|ccr}
\toprule
 & \multicolumn{3}{c}{Time-complexity}  \\
& ORFs & SimRFs & SimRFs+ \\
\midrule
Regular    & $\mathcal{O}(d^3)$ & $\mathcal{O}(d^3)$ & $\mathcal{O}(d^3)$  \\
Fast    & $\mathcal{O}(d \log d)$ & $\mathcal{O}(d \log d)$ & $\mathcal{O}(d^3)$  \\
\bottomrule
\end{tabular}
\end{small}
\end{center}
\vskip -0.1in
\end{table}
\vspace{-4mm}
For all regular schemes, the space complexity to store $\mathbf{R}$ is $\mathcal{O}(d^2)$. For fast ORFs and fast SimRFs, the space complexity becomes  $\mathcal{O}(d)$ because we no longer need to explicitly store $\widetilde{\mathbf{R}}$, just the $d$ weights $\{w_i\}$ from $\chi_d$. But the space complexity of fast SimRFs+ is still  $\mathcal{O}(d^2)$ since all vectors must be stored during the optimisation step.

It is clear that \textbf{SimRFs are essentially equal in computational cost to ORFs}, and in Sec. \ref{experiments} we will see that they often perform substantially better in downstream tasks. Meanwhile, SimRFs+ are mostly of academic interest.
\section{Experiments} 
\label{experiments}
Here we report the outcomes of an extensive empirical evaluation of SimRFs for PRFs, demonstrating their superiority over IIDRFs and ORFs in a variety of settings. Technical details are reported in Appendix \ref{exp_appendix}. The section is organised as follows: (a) in Sec. \ref{var_comp} we plot the derived MSE expressions for IIDRFs, ORFs and SimRFs; (b) in Sec. \ref{sec:frobnorm} we verify that SimRFs permit higher-quality kernel matrix approximation by considering the Frobenius norm of the difference between the true and approximated Gram matrices; (c) in Sec. \ref{nonparam_class} we compare the performance of the different RF mechanisms on nonparametric classification tasks using kernel regression; (d) in Sec. \ref{exp_performers} we compare the RF mechanisms for approximation of the attention module in vision Performer-Transformers.

\subsection{Comparison of MSE Between RF Mechanisms} \label{var_comp}

\begin{figure}[H]
\vspace{-3mm}
\centering
  \includegraphics{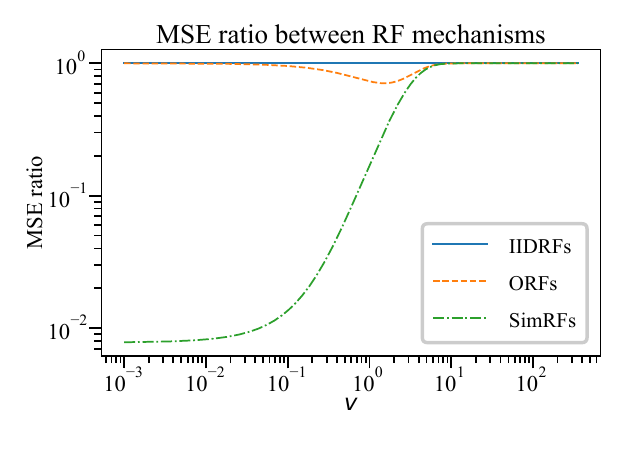}
  \vspace{-3mm}
  \caption{Analytic form the the MSE ratio of the PRF kernel estimator $\widehat{K}$ for different couplings, plotted as a function of $v=\|\boldsymbol{x}+\boldsymbol{y}\|_2$. Smaller values indicate lower MSE and are hence better. SimRFs always perform the best, followed by ORFs then IIDRFs. The size of the improvement depends on the data; it is bigger at smaller $v$.}
\vspace{-5mm}
\label{var_comp_fig}
\end{figure}

We begin by plotting the MSE of the PRF estimator $\widehat{K}$ with IIDRFs, ORFs and SimRFs, given by Eq. \ref{eq:def_rf_conformity} with the RF-confirmities \ref{eq:iidrf_conformity}, \ref{eq:orf_conformity} and \ref{eq:simrf_conformity}. We note that the \emph{ratio} of the MSE of any pair of RF mechanisms only depends in the data $\boldsymbol{x},\boldsymbol{y}$ via $v=\| \boldsymbol{x}+\boldsymbol{y}\|_2$, so it is natural to plot $\text{MSE}_\text{ORF}/\text{MSE}_\text{IIDRF}$ and $\text{MSE}_\text{SimRF}/\text{MSE}_\text{IIDRF}$ as a function of $v$ -- see Fig. \ref{var_comp_fig}. We take $d=64$ which is standard in Transformer applications. 

SimRFs always outperform ORFs and IIDRFs, but the size of the improvement depends sensitively on the data. SimRFs are particularly effective compared to ORFs and IIDRFs when estimating kernel evaluations at small $v$. This can be understood from their respective Taylor expansions. For both IIDRFs and ORFs, the MSE goes as $\textrm{MSE}_\textrm{IIDRF,ORF}=v^2 + \mathcal{O}(v^4)$. Meanwhile, for SimRFs, $\textrm{MSE}_\textrm{SimRF}=v^2 \left(1 - \frac{\sqrt{\pi} \Gamma(d+1)\Gamma(\frac{d}{2}+\frac{1}{2})}{\Gamma(\frac{d}{2}) \Gamma(\frac{d}{2}+1)^2 2^d} \right) + \mathcal{O}(v^4)$. For $d=64$, the SimRF $v^2$ prefactor evaluates to $0.0078$ which is manifestly substantially smaller than $1$. 

\subsection{Quality of Gram Matrix Approximation}\label{sec:frobnorm}
\begin{figure}[H]
\vspace{-3mm}
\centering
  \includegraphics{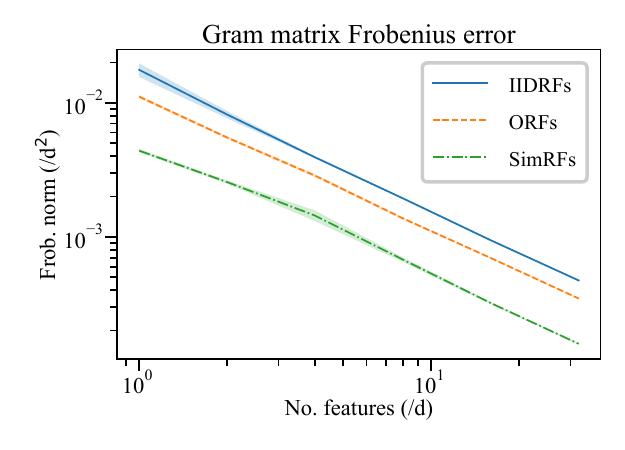
}
  \vspace{-3mm}
  \caption{Frobenius norm between the true and approximated Gram matrices (lower is better) using different RF mechanisms and a different number of random features. More features give a better approximation, and SimRFs consistently outperform ORFs and IIDRFs. The data is of dimensionality $d=64$ and we take $N=64$ points, generated normally with $\sigma=0.1$. The shading gives one standard deviation on estimates of the mean.}
\vspace{-3mm}
\label{fig:frobnorm}
\end{figure}

Another straightforward task is to directly compare the quality approximation of the Gram matrix $\widehat{\mathbf{K}}$ with the different RF mechanisms. We can quantify this using the Frobenius norm between the exact and approximated matrices $\sum_{i=1}^N \sum_{j=1}^N (\mathbf{K}_{ij} - \widehat{\mathbf{K}}_{ij})^2$, where $\mathbf{K}_{ij} \overset{\textrm{def}}{=} K_\textrm{gauss}(\boldsymbol{x}_i,\boldsymbol{x}_j)$  and $\widehat{\mathbf{K}}_{ij}$ is the corresponding low-rank decomposition. For demonstration purposes, we randomly generate $N=64$ data points of dimensionality $d=64$ according to the distribution $\boldsymbol{x}_i \sim \mathcal{N}(0,\sigma^ 2 \mathbf{I}_d)$. We take $\sigma=0.1$. Fig. \ref{fig:frobnorm} shows the results; the quality of Gram matrix approximation improves with the number of features, and is better with SimRFs than ORFs and IIDRFs. 
\subsection{Nonparametric Classification Using Kernel Regression} \label{nonparam_class}
Here we demonstrate how reduced kernel estimator MSE translates to better performance in downstream classification tasks. We use $8$ different datasets retrieved from the UCI Machine Learning Repository \cite{Dua:2019}, each consisting of $L$ training data $\{(\boldsymbol{x},\boldsymbol{y})\}$ and test data $\{(\boldsymbol{x'},\boldsymbol{y'})\}$. The objects are $d$-dimensional vectors $\boldsymbol{x},\boldsymbol{x}' \in \mathbb{R}^d$ and their labels are one-hot encoded  $\boldsymbol{y},\boldsymbol{y}' \in \mathbb{R}^n$. We predict the label distribution of a test object using kernel regression with the Gaussian kernel, $\boldsymbol{y}'_\text{pred} = \sum_{i=1}^L K(\sigma \boldsymbol{x}', \sigma \boldsymbol{x}^{(i)}) \boldsymbol{y}^{(i)}/\sum_{i=1}^L K(\sigma \boldsymbol{x}', \sigma \boldsymbol{x}^{(i)})$. We then predict a class by taking the greatest argument of $\boldsymbol{y}'_\text{pred}$. 
\begin{figure}
\centering
  \includegraphics
  {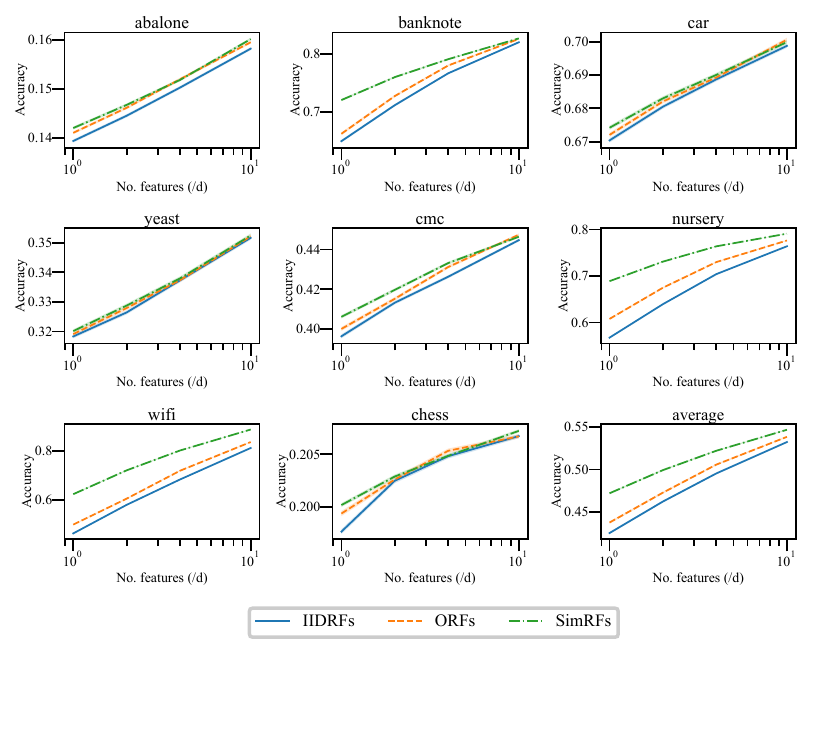}
  \vspace{-15mm}
  \caption{Nonparametric classification using kernel regression for a variety of datasets \cite{Dua:2019,nash1994population,Dua:2019_banknote,bohanec1988knowledge,horton1996probabilistic,lim2000comparison,olave1989application,Dua:2019_chess}, where the Gaussian kernel is approximated with different RFs. Plots show mean classification accuracy vs the number of random features used to approximate the kernel (/$d$, the dimensionality of the objects $\boldsymbol{x}$). Shading gives the standard deviation on the estimates of the mean. SimRFs consistently perform best.}
\vspace{-5mm}  
\label{nonparam_results}
\end{figure}We measure accuracy by the proportion of correct label predictions across the test-set. The $\sigma>0$ hyperparameter is tuned for good PRF performance on a validation dataset; see Appendix \ref{sigmachoice} for detailed discussion. Fig. \ref{nonparam_results} presents the results, plotting classification accuracy against the number of random features used.
The size of the benefit accrued from using SimRFs depends on the data (as we noted in Sec. \ref{var_comp}) and in the limit of large $m$ performance tends towards the exact kernel result. SimRFs consistently perform best.

\subsubsection{SimRFs+ for Nonparametric Classification}
Table \ref{tab:splus_table} compares the classification accuracies achieved with SimRFs and SimRFs+ on the task detailed above, using $m=d$ random features. As suggested in Sec. \ref{sec:algorithm} (see in particular Fig. \ref{corr_inset_main}), SimRFs are already close to optimal and any gain provided by using SimRFs+ is marginal. Moreover, improvements tend to occur where $v$ is small so truncating the objective series expansion at $k=1$ is reasonable.
\vspace{-5mm}
\begin{table}[H]
\caption{
Classification accuracies from kernel regression with SimRFs and SimRFs+, using random features of length $m=d$. $\bar{v}$ records the mean ($\sigma$-scaled) value of $v$ in each dataset. Note that both variants substantially outperform ORFs on every dataset.}
\label{tab:splus_table}\vspace{-0mm}
\begin{center}
\begin{small}
\begin{tabular}{l|c|ccr}
\toprule
Data set & $\bar{v}$ & \multicolumn{2}{c}{Classification accuracy}  \\
& & SimRFs & SimRFs+ \\
\midrule
$\mathrm{abalone}$    & 1.7 & \textbf{0.1421$\pm$0.0002} & \textbf{0.1419$\pm$0.0002}  \\
$\mathrm{banknote}$ & 2.6&  \textbf{0.7229$\pm$0.0012} & 0.7132$\pm$0.0012 \\
$\mathrm{car}$    & 5.0 & \textbf{0.6754$\pm$0.0004} & \textbf{0.6751$\pm$0.0004}\\
$\mathrm{yeast}$    & 3.1 & \textbf{0.3202$\pm$0.0004} & \textbf{0.3208$\pm$0.0004}   \\
$\mathrm{cmc}$     & 2.0 & 0.4047$\pm$0.0005 & \textbf{0.4065$\pm$0.0005} \\
$\mathrm{nursery}$      & 1.4 & 0.6874$\pm$0.0005 & \textbf{0.6917$\pm$0.0004} \\
$\mathrm{wifi}$      &0.8 & 0.6314$\pm$0.0018 & \textbf{0.6473$\pm$0.0018}       \\
$\mathrm{chess}$   & 2.3 & \textbf{0.2000$\pm$0.0001} & \textbf{0.2000$\pm$0.0001} \\
\bottomrule
\end{tabular}
\end{small}
\end{center}
\vskip -0.1in
\end{table}

\subsection{SimRFs-Performers: Scalable Attention for Transformers}\label{exp_performers}

 PRFs were first introduced in \cite{choromanski2020rethinking} in order to accurately approximate the softmax attention module of Transformers -- an architecture coined the \emph{Performer}. This technique for kernelising the attention mechanism, which identifies complex dependencies between the elements of an input sequence, permits linear (c.f. quadratic) space- and time-complexity without assuming restrictive priors such as sparsity and low-rankness. Performers offer competitive results across a range of tasks \cite{lra-paper}, including vision modeling \cite{perf-vit-1, perf-vit-2} and speech \cite{perf-speech}.

Since Performers apply the ORF variant of PRFs, it is natural to expect that the SimRFs mechanism, which gives provably lower kernel estimator MSE, will be more effective. We refer to this architecture as the \emph{SimRFs-Performer}, and show that it outperforms the regular ORFs-Performer.  

We focus on the `performised' versions of Vision Transformers (ViTs) \cite{vits}
and consider four datasets: (a) \textrm{ImageNet2012} \cite{imagenet} (1K classes, 1.2M training images, 100K test set); (b) \textrm{Fashion-MNIST} \cite{fashionmnist} (10 classes, 60K training images, 10K test set); (c) \textrm{I\_naturalist2021} \cite{inaturalist} (10K classes, 2.7M training images, 500K test set) and (d) \textrm{Places365} \cite{places} (365 classes, 1.8M training images, 328K test set). These are often used to benchmark ViTs.

In all four experiments, we use a ViT with 12 layers, 12 heads, mlp\_dim equal to 3072, a dropout rate of 0.1 and no attention dropout. We use the $\mathrm{adam}$ optimiser with weight decay equal to 0.1 and batch size $\mathrm{bs}=4096$, trained for 300 epochs on the $\mathrm{TPU}$ architecture. We apply 130 random vectors to approximate the softmax attention kernel with PRFs, testing both the ORF and SimRF coupling mechanisms.

\begin{figure}
\centering
  \includegraphics[width = 0.99\linewidth]{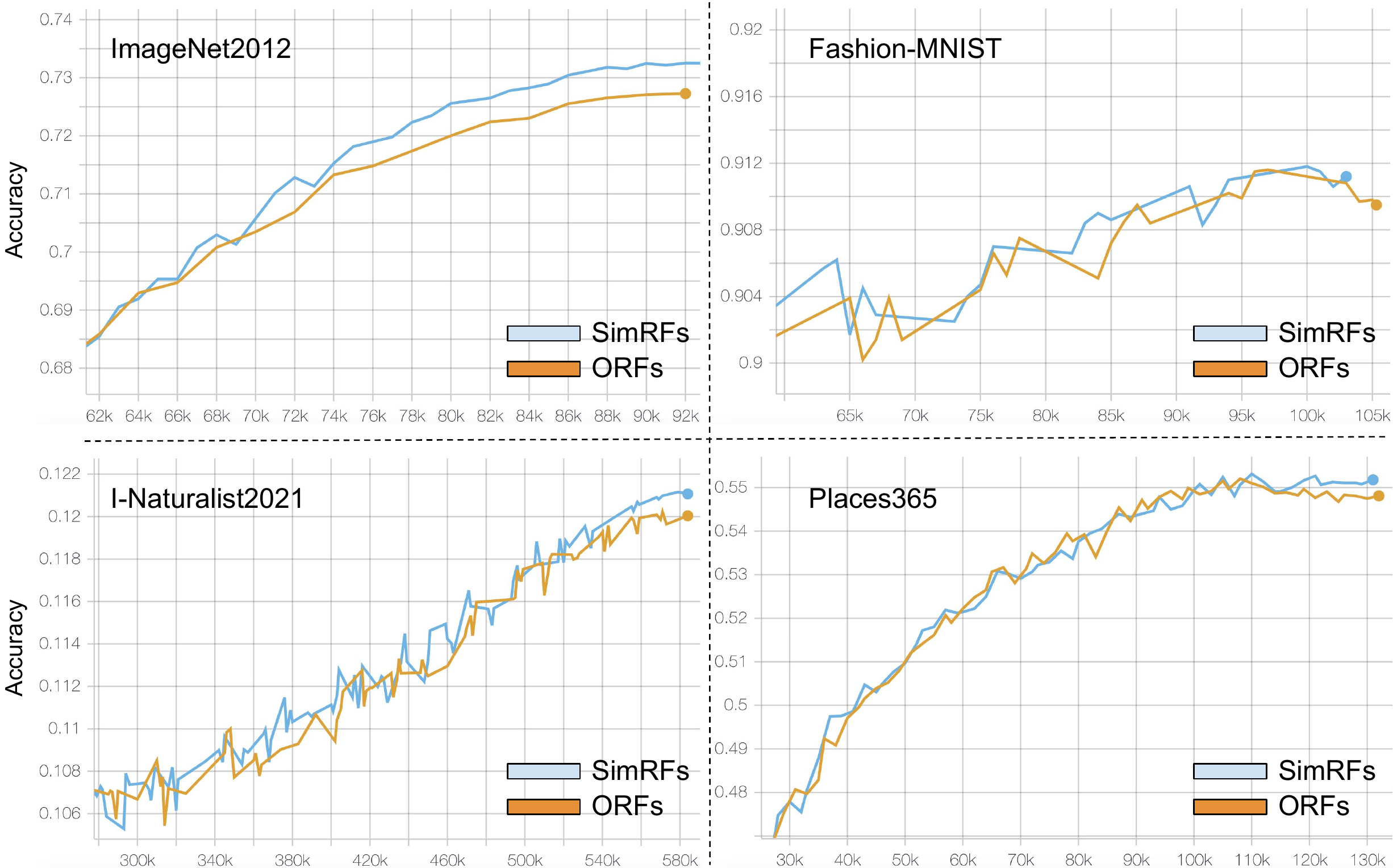}
  \vspace{-3mm}
  \caption{\small{Accuracy comparison (higher is better) of the SimRFs-Performer and the regular ORFs-Performer. Tests are on four image classification tasks: (a) ImageNet2012, (b) Fashion-MNIST, (c) I-Naturalist2021, (d) Places365}. $x$-axis is training epochs.}
\vspace{-5mm}
\label{performer_fig}
\end{figure}

The results, comparing ORFs and SimRFs for approximating attention, are presented in Fig. \ref{performer_fig}. The SimRFs-Performer often achieves gains over the regular ORFs-Performer -- and is certainly never worse -- for no observable extra cost. The exact difference depends on the data distribution (see. Sec. \ref{var_comp}) and the importance of MSE reduction for that particular task; if some other factor is bottlenecking Performer accuracy, then improving the approximation of the attention matrix cannot provide gains. Nonetheless, for some of the tested datasets the difference is substantial: for instance, on $\mathrm{ImageNet2012}$, which is frequently used to benchmark new Transformer variants, the SimRFs-Performer saturates at an accuracy which is greater than the regular ORFs-Performer by $\textbf{0.5\%}$. It is remarkable that such a large gain can be accrued with a single drop-in matrix multiplication at no observable computational cost, without any architectural or ViT-specific changes. 

\section{Conclusion}
We have introduced \emph{Simplex Random Features} (SimRFs), a new mechanism for unbiased approximation of the Gaussian and softmax kernels. By correlating the directions of random vectors in the ensemble, we access lower kernel estimator MSE than the previously predominant \emph{Orthogonal Random Features} (ORFs): a fact we have verified both theoretically and empirically via extensive experiments. We have shown that the suppressed MSE of SimRFs compared to ORFs often permits better performance in downstream applications, including in nonparametric classification and scalable Transformer training. However, the size of the gain depends on the data distribution and whether the quality of kernel approximation is currently bottlenecking model performance. We have proved that SimRFs constitute the best weight-independent geometrically-coupled PRF mechanism, with further marginal improvements available in some regimes from a weight-dependent SimRFs+ variant. Finally, through our detailed quantitative analysis of the different RF mechanisms, we have derived novel closed-form results for ORFs, precisely formalising qualitative and asymptotic findings previously reported in the literature. 

\section{Relative Contributions and Acknowledgements} 
IR developed the SimRF and SimRF+ mechanisms, proved all theoretical results, and ran the pointwise kernel evaluation, Frobenius norm and nonparametric classification experiments. KC designed and ran the Performer experiments, and was crucially involved in all aspects of the work throughout. AW and VL provided helpful discussion and feedback on drafts. 

IR acknowledges support from a Trinity College External Studentship. VL acknowledges support from the Cambridge Trust and DeepMind. AW acknowledges support from a Turing AI Fellowship under grant EP/V025279/1 and the Leverhulme Trust via CFI.

\newpage

\bibliographystyle{icml2023}

\newpage
\appendix
\onecolumn
\section{Supplementary Proofs and Discussion}
\label{app:all_proofs_appendix}
In this appendix we provide further discussion and proofs of results stated in the main text.
\subsection{Proof of Theorem \ref{thm:rf-conformity} (MSE Depends on RF-conformity)} \label{app:conformity}
We begin by deriving a form for the kernel estimator MSE in the PRF setting, showing how it depends upon the so-called  \emph{RF-conformity} defined in Eq. \ref{eq:def_rf_conformity}.

From the definitions in Eq. \ref{eq:def_gauss_kern} and Eq. \ref{eq:def_prf}, it follows that 

\begin{equation}
    \widehat{K} =\phi(\boldsymbol{x})^{\top} \phi(\boldsymbol{y}) \\= \frac{e^{-x^2-y^2}}{m}\sum_{i=1}^m e^{\boldsymbol{w}_i^{\top}(\boldsymbol{x}+\boldsymbol{y})} = \frac{e^{-x^2-y^2}}{m} \sum_{i=1}^m b_i
\end{equation}
where $m$ is the number of random features and $\boldsymbol{x},\boldsymbol{y} \in \mathbb{R}^d$. We introduced $b_i$, where
\begin{equation}
b_i \overset{\mathrm{def}}{=} e^{\boldsymbol{w}_i^{\top} \boldsymbol{v}},
\end{equation}
with $\boldsymbol{v} = \boldsymbol{x}+\boldsymbol{y} \in \mathbb{R}^d$. Here, $i=1,...,m$ enumerates the random features. It is straightforward to show that this is an unbiased estimator of the Gaussian kernel $K(\boldsymbol{x}, \boldsymbol{y}) = \exp(-\frac{\|\boldsymbol{x}-\boldsymbol{y}\|_2^2}{2})$ when $\boldsymbol{w}_i$ are sampled from $\mathcal{N}(0,\mathbf{I}_d)$; in particular, we find that $\mathbb{E}(b_i) = e^\frac{v^2}{2}$. After some algebra, we can also show that
\begin{equation}
    \text{MSE}(\widehat{K})= \frac{e^{-2x^2-2y^2}}{m}\left ((e^{2v^2}- e^{v^2})+  (m-1)(\frac{1}{m(m-1)}\sum_i\sum_{i\neq j}\mathbb{E}[b_ib_j]-e^{v^2}) \right).
\label{prf_variance_2}
\end{equation}
Now consider the correlation term $\frac{1}{m(m-1)}\sum_i\sum_{i\neq j}\mathbb{E}[b_ib_j] = \frac{1}{m(m-1)}\sum_i\sum_{i\neq j}\mathbb{E}[e^{(\boldsymbol{w}_i + \boldsymbol{w}_j)^{\top} \boldsymbol{v}}]$ more carefully. Evidently, we care about the probability distribution over the random variable $\boldsymbol{w}_i + \boldsymbol{w}_j$, denoted compactly by $\boldsymbol{w}_{ij}$. For all couplings we consider, the random vectors $\boldsymbol{w}_i$ and $\boldsymbol{w}_j$ are marginally isotropic (a necessary condition to be marginally Gaussian) and their resultant $\boldsymbol{w}_{ij}$ will also be marginally isotropic. This permits us to rewrite the expectation value using the \emph{Hankel transform} \cite{arizona_hankel}:
\begin{equation} \label{eq:prf_hankel}
\begin{multlined}
\mathbb{E}[e^{\boldsymbol{w}_{ij}^{\top} \boldsymbol{v}}] = \int_{\mathbb{R}_d} \text{d}^d\boldsymbol{w}_{ij} p(\boldsymbol{w}_{ij}) e^{\boldsymbol{w}_{ij}^{\top}\boldsymbol{v}}
\\ = \Gamma(d/2) 2^{\frac{d}{2}-1} \int_0^\infty \text{d}w_{ij} p(w_{ij}) (iw_{ij}v)^{1-\frac{d}{2}} J_{\frac{d}{2}-1} (iw_{ij}v)
\end{multlined}
\end{equation}
where $J_{\frac{d}{2}-1}$ is a Bessel function of the first kind\footnote{In fact, given the purely imaginary argument, the function $I_\alpha(x) = i^{-\alpha} J_\alpha(ix)$ is referred to as the \emph{modified} Bessel function.}. Importantly, we are integrating over a single variable: the norm of the resultant random vector, $w_{ij} = \|\boldsymbol{w}_i + \boldsymbol{w}_j \|_2$. The probability distribution $p(w_{ij})$ will depend on whether the random vectors are i.i.d. or exhibit geometrical coupling, even though the marginal distributions are identical (Gaussian) in every case. 

Recalling the Taylor expansion \begin{equation} \label{eq:besselboy}
    J_\alpha(z) = \sum_{k=0}^\infty \frac{(-1)^k}{k! \Gamma(k + \alpha +1)} \left(\frac{z}{2}\right)^{2k + \alpha},
\end{equation}
we can rewrite the correlation term as
\begin{equation}
\mathbb{E}[b_ib_j] = \Gamma \left(\frac{d}{2}\right) \mathbb{E}_{w_{ij}}  \left(\sum_{k=0}^\infty \frac{v^{2k} w_{ij} ^{2k}}{2^{2k} k! \Gamma(k+\frac{d}{2})}\right).
\end{equation}
Inserting this into Eq. \ref{prf_variance_2}, this immediately yields the important result:
\begin{equation}
    \text{MSE}(\widehat{K})= \frac{e^{-2x^2-2y^2}}{m}\left ((e^{2v^2}- e^{v^2})+  (m-1)(\rho(\boldsymbol{x},\boldsymbol{y}) - e^{v^2}) \right),
\label{prf_variance_3}
\end{equation}
where we defined the \emph{RF-conformity}
 \begin{equation} \label{ensemble_exp_app}
 \rho(\boldsymbol{x}, \boldsymbol{y})\overset{\mathrm{def}}{=}
\frac{\Gamma(\frac{d}{2})}{m(m-1)} \sum_{i} \sum_{j\neq i} \mathbb{E}_{w_{ij}}\left( \sum_{k=0}^\infty \frac{v^{2k} w_{ij} ^{2k}}{2^{2k} k! \Gamma(k+\frac{d}{2})} \right),
\end{equation}
as in Eq. \ref{eq:def_rf_conformity} of the main text. Summations run from $i=1$ to $m$, the number of random features. The MSE is manifestly an increasing function of $\rho(\boldsymbol{x},\boldsymbol{y})$, which itself depends sensitively on the any correlations induced between random vectors via $p(w_{ij})$. It is clear that any coupling mechanisms that reduce values of $w_{ij}=\|\boldsymbol{w}_i+\boldsymbol{w}_j\|_2$, e.g. by conditioning that random vectors point away from one another, will suppress $\rho(\boldsymbol{x},\boldsymbol{y})$. The RF-conformity will form a core consideration in the discussion that follows.

\subsection{Proof of Theorem \ref{thm:simplex_optimal} (SimRFs Optimal for Weight-Independent Geometrical Coupling)}\label{app:equal_weights}
Here, we prove the central result that, supposing the weights $w_i = \|\boldsymbol{w_i}\|_2, i=1,...,d$ are i.i.d. (in our case from $\chi_d$), SimRFs constitute the best possible weight-independent geometrical coupling scheme. Recall again that by `weight-independent' we mean that vector directions $\{\hat{\boldsymbol{w}_i}\}$ are independent of norms $\{w_i\}$, though directions can still be correlated among themselves. Our choice of geometrical coupling will not depend on each particular draw of norms; we just use the fact that all $w_i$ are identically distributed.

We begin by proving the following simpler auxiliary lemma. 

\begin{lemma}[SimRFs optimal for equal norms] \label{lemma:equal_weights}
Suppose that, instead of being sampled from a $\chi_d$ distribution, we condition that $\boldsymbol{w}_{i} \in \mathbb{R}^{d}$ for $i=1,...,d$ all have equal lengths $w$. Then $\rho(\boldsymbol{x},\boldsymbol{y})$ is minimised when the ensemble exhibits simplex geometrical coupling. 
\end{lemma}

\underline{Proof:}
given the set of vector norms $w_i = w$ with $i=1,...,d$, we would like to know how to choose the angles $\theta_{ij}$ subtended between each pair $\boldsymbol{w}_i$ and $\boldsymbol{w}_j$ to minimise the RF-conformity $\rho(\boldsymbol{x},\boldsymbol{y})$. It is immediately obvious that we should choose $\theta_{ij}$ deterministically rather than probabilistically, because assigning probability mass to suboptimal configurations will always increase the expectation value (that is, $p(w_{ij}|w_i,w_j=w) = \delta(w_{ij} - \sqrt{2w^2(1+\cos\theta_{ij})})$, with $\delta$ the delta function). So the task is to choose $\{\theta_{ij}\}$ to minimise
 \begin{equation}
 \rho(\boldsymbol{x}, \boldsymbol{y})=
\frac{\Gamma(\frac{d}{2})}{m(m-1)} \sum_{i} \sum_{j\neq i}  \sum_{k=0}^\infty \frac{v^{2k} w^{2k}}{2^{2k} k! \Gamma(k+\frac{d}{2})}\| \widehat{\boldsymbol{w}}_i + \widehat{\boldsymbol{w}}_j \|_2 ^{2k} = \sum_{i} \sum_{j \neq i} f(\| \widehat{\boldsymbol{w}}_i + \widehat{\boldsymbol{w}}_j \|_2^2)
\end{equation}
where we defined the increasing convex function
\begin{equation}
f(\| \widehat{\boldsymbol{w}}_i + \widehat{\boldsymbol{w}}_j \|_2^2 ) \overset{\mathrm{def}}{=} \frac{\Gamma(\frac{d}{2})}{m(m-1)} \sum_{k=0}^\infty \frac{v^{2k} w^{2k}}{2^{2k} k! \Gamma(k+\frac{d}{2})}\| \widehat{\boldsymbol{w}}_i + \widehat{\boldsymbol{w}}_j \|_2 ^{2k}.
\end{equation}
It follows from Jensen's inequality that
\begin{equation}
\sum_{i} \sum_{j \neq i} f(\|\widehat{\boldsymbol{w}}_i + \widehat{\boldsymbol{w}}_j\|_2^2) \geq m(m-1) f\left(\frac{\sum_{i} \sum_{j \neq i} \| \widehat{\boldsymbol{w}}_i + \widehat{\boldsymbol{w}}_j \|_2^2 }{m(m-1)}\right)
\end{equation}
with equality when $\| \widehat{\boldsymbol{w}}_i + \widehat{\boldsymbol{w}}_j \|_2$ is identical for every $i,j$, i.e. all random vectors subtend equal angles. Since $f$ is increasing,

\begin{equation}
\begin{multlined}
\sum_i \sum_{j \neq i} \| \widehat{\boldsymbol{w}}_i + \widehat{\boldsymbol{w}}_j \|_2^2 = \sum_i \sum_{j \neq i} 2 + 2\widehat{\boldsymbol{w}}_i^\top \widehat{\boldsymbol{w}}_j
\\ = 2m(m-1) + 2 \sum_i \widehat{\boldsymbol{w}}_i^\top  \sum_{j \neq i} \widehat{\boldsymbol{w}}_j
\\ =  2m(m-2) + 2( \sum_i \widehat{\boldsymbol{w}}_i)^\top ( \sum_{j} \widehat{\boldsymbol{w}}_j)
\\ =  2m(m-2) + 2 \|\sum_i \widehat{\boldsymbol{w}}_i\|_2^2 \geq 2m(m-2)
\end{multlined}
\end{equation}
with equality achieved when $\sum_{i} \widehat{\boldsymbol{w}}_i=0$. Therefore, 
\begin{equation}
\rho (\boldsymbol{x},\boldsymbol{y}) = \sum_{i} \sum_{j \neq i} f(\|\widehat{\boldsymbol{w}}_i + \widehat{\boldsymbol{w}}_j\|_2^2) \geq m(m-1) f\left(\frac{2(m-2)}{m-1}\right).
\end{equation}
This shows that the conformity is minimised when we have that i) all vectors $\widehat{\boldsymbol{w}}_i$ subtend equal angles, and ii) $\sum_{i} \widehat{\boldsymbol{w}}_i=0$. This is nothing other than the geometry of a $d-1$-dimensional simplex embedded in $d$-dimensional space, as described by the basis vectors defined in Eq. \ref{eq:simplex_vectors}. \qed

Armed with the result of Lemma \ref{lemma:equal_weights}, we now consider the more general setting where $\{w_i\}$ are i.i.d. random variables but draws are not generically identical. 

We begin with the observation that, if random variables $w_1,...,w_d$ are i.i.d., the joint distribution $p(w_1,w_2,...,w_d)=p(w_1)\cdot p(w_2)\cdot ... \cdot p(w_d)$ is invariant under permutation of $w_i$. This is because the joint distribution factorises into $d$ identical functions, though more general joint distributions with this property exist. Intuitively, for every given draw of weights $\{w_1,...,w_d\}$, there are $d!-1$ other draws of equal probability given by the permutations $\{w_{P_1},...,w_{P_d}\}$ where $P \in S_d$, the symmetric group on $d$ letters. Therefore, the RF conformity can be expressed as
\begin{equation}  \label{eq:perm_sum_warmup}
\begin{multlined}
 \rho(\boldsymbol{x}, \boldsymbol{y})=
\frac{\Gamma(\frac{d}{2})}{m(m-1)}  \int \mathrm{d}w_1 \mathrm{d}w_2 ... \mathrm{d}w_d p(w_1,...,w_d) \sum_{i} \sum_{j \neq i} \sum_{k=0}^\infty \frac{v^{2k} w_{ij} ^{2k}}{2^{2k} k! \Gamma(k+\frac{d}{2})} 
\\ =\frac{\Gamma(\frac{d}{2})}{m(m-1)}  \int \mathrm{d}w_1 \mathrm{d}w_2 ... \mathrm{d}w_d \frac{p(w_1,...,w_d)}{d!} \\ \cdot \sum_{P \in S_d} \sum_{i} \sum_{j \neq i} \sum_{k=0}^\infty \frac{v^{2k}}{2^{2k} k! \Gamma(k+\frac{d}{2})}\left(w_{P_i}^2 \hat{\boldsymbol{w}}_i^\top \hat{\boldsymbol{w}}_i + w_{P_j}^2 \hat{\boldsymbol{w}}_j^\top \hat{\boldsymbol{w}}_j + 2 w_{P_i}w_{P_j} \hat{\boldsymbol{w}}_i^\top \hat{\boldsymbol{w}}_j  \right)^k
\end{multlined}
\end{equation} 
where we wrote $p(w_1,...,w_d)=\frac{1}{d!} \sum_{P\in S_n}p(w_{P_1} ,w_{P_2},...,w_{P_d})$ then relabelled the integration variables. Here, we have permuted the random vector norms $w_i$ but \emph{not} the directions $\hat{\boldsymbol{w}}_i$. We would like to obtain the geometry $\{\hat{\boldsymbol{w}}_i\}$ that minimises $\rho(\boldsymbol{x},\boldsymbol{y})$, subject to the condition that the normalisations of the unit vectors $\hat{\boldsymbol{w}}_i^\top \hat{\boldsymbol{w}}_i=1$ are fixed. Since the integrand is nonnegative everywhere, we minimise the sum in the final line of Eq. \ref{eq:perm_sum_warmup}, namely
\begin{equation}
\begin{multlined}
\sum_{P \in S_d} \sum_{i} \sum_{j \neq i}   f\left(w_{P_i}^2 + w_{P_j}^2 + 2 w_{P_i}w_{P_j} \hat{\boldsymbol{w}}_i^\top \hat{\boldsymbol{w}}_j  \right)
\\ = \sum_{P \in S_d} \sum_{i} \sum_{j \neq i} \sum_{k=0}^\infty \frac{v^{2k}}{2^{2k} k! \Gamma(k+\frac{d}{2})}\left(w_{P_i}^2 + w_{P_j}^2 + 2 w_{P_i}w_{P_j} \hat{\boldsymbol{w}}_i^\top \hat{\boldsymbol{w}}_j  \right)^k
\end{multlined}
\end{equation}
where $f$ is once again convex and positive definite. Relabelling summation variables then using Jensen's inequality, we can write this as 
\begin{equation}
\sum_{P \in S_d} \sum_{i} \sum_{j \neq i}   f\left(w_{i}^2 + w_{j}^2 + 2 w_{i}w_{j} \hat{\boldsymbol{w}}_{P_i}^\top \hat{\boldsymbol{w}}_{P_j}  \right)
\geq d! \sum_{i} \sum_{j \neq i}   f\left(\frac{\sum_{P \in S_d}w_{i}^2 + w_{j}^2 + 2 w_{i}w_{j} \hat{\boldsymbol{w}}_{P_i}^\top \hat{\boldsymbol{w}}_{P_j}}{d!}  \right)
\end{equation}
with equality when $\hat{\boldsymbol{w}}_{P_i}^\top \hat{\boldsymbol{w}}_{P_j}$ is identical for every permutation -- that is, when all the random vectors subtend identical angles. With this in mind, we write the Lagrangian as
\begin{equation}
\mathcal{L} = \sum_{P \in S_d} \sum_{i} \sum_{j \neq i} \sum_{k=0}^\infty \frac{v^{2k}}{2^{2k} k! \Gamma(k+\frac{d}{2})}\left(w_{P_i}^2 \hat{\boldsymbol{w}}_i^\top \hat{\boldsymbol{w}}_i + w_{P_j}^2 \hat{\boldsymbol{w}}_j^\top \hat{\boldsymbol{w}}_j + 2 w_{P_i}w_{P_j} \hat{\boldsymbol{w}}_i^\top \hat{\boldsymbol{w}}_j  \right)^k - \sum_i \lambda_i (\hat{\boldsymbol{w}}_i^\top \hat{\boldsymbol{w}}_i - 1).
\end{equation}
Differentiating wrt $\hat{\boldsymbol{w}}_i$,
\begin{equation} \label{eq:permuted_opt_1}
\sum_{P \in S_d} \sum_{j \neq i} \sum_{k=0}^\infty \frac{v^{2k}k w_{P_i P_j}^{2k-2}}{2^{2k} k! \Gamma(k+\frac{d}{2})}\left(w_{P_i}^2 \hat{\boldsymbol{w}}_i +  w_{P_i}w_{P_j} \hat{\boldsymbol{w}}_j  \right)  - \lambda_i \boldsymbol{w}_i
= 0 \hspace{5ex} i=1,...,d. \end{equation}
where we used that $\hat{\boldsymbol{w}}_{P_i}^\top \hat{\boldsymbol{w}}_{P_j} = \hat{\boldsymbol{w}}_{i}^\top \hat{\boldsymbol{w}}_{j}$ to take
\begin{equation}
w_{P_i}^2 \hat{\boldsymbol{w}}_i^\top \hat{\boldsymbol{w}}_i + w_{P_j}^2 \hat{\boldsymbol{w}}_j^\top \hat{\boldsymbol{w}}_j + 2 w_{P_i}w_{P_j} \hat{\boldsymbol{w}}_i^\top \hat{\boldsymbol{w}}_j  = w_{P_i P_j} =  w_{P_i}^2 \hat{\boldsymbol{w}}_{P_i}^\top \hat{\boldsymbol{w}}_{P_i} + w_{P_j}^2 \hat{\boldsymbol{w}}_{P_j}^\top \hat{\boldsymbol{w}}_{P_j} + 2 w_{P_i}w_{P_j} \hat{\boldsymbol{w}}_{P_i}^\top \hat{\boldsymbol{w}}_{P_j}.
\end{equation}

Eq. \ref{eq:permuted_opt_1} implies that 
\begin{equation}
\hat{\boldsymbol{w}}_i \propto -  \sum_{j \neq i} \sum_{k=0}^\infty \frac{v^{2k}k}{2^{2k} k! \Gamma(k+\frac{d}{2})} \left ( \sum_{P \in S_d} w_{P_i P_j}^{2k-2} w_{P_i}w_{P_j} \right ) \hat{\boldsymbol{w}}_j  
 \hspace{5ex} i=1,...,d
\end{equation}
with the proportionality constant fixed by the normalisation of $\hat{\boldsymbol{w}}_i$. Crucially, since we are summing over all permutations $S_d$ of the $d$ labels, the term in parentheses $\left( \sum_{P \in S_d} w_{P_i P_j}^{2k-2} w_{P_i}w_{P_j}\right)$ is identical for every $i,j$. This immediately implies that 
\begin{equation}
\hat{\boldsymbol{w}}_i \propto -  \sum_{j \neq i} \hat{\boldsymbol{w}}_j   \hspace{5ex} i=1,...,d.
\end{equation}
Subject to the further constraint that all $\hat{\boldsymbol{w}}_i$ subtend equal angles, this is uniquely given by the simplex geometry described by the basis vectors in Eq. \ref{eq:simplex_vectors}. That is, supposing the vector norms $w_i$ are i.i.d. and that the geometrical coupling is weight-independent, SimRFs give the lowest possible MSE in the PRF setting. \qed

An intuitive explanation of this result is as follows. In Lemma \ref{lemma:equal_weights}, we observed that SimRFs are optimal if all vector norms $w_i$ are equal. Supposing norms are not equal but are identically distributed, any geometrical coupling scheme that is better for some particular draw of norms $\{w_i\}$ will be worse for some of the (equally probable) label permutations $\{w_{P_i}\}, P \in S_d$. The effect of summing over all the permutations is the same as collapsing all the distributions over $w_i$ to a single, identical value.

\subsection{Derivation of Eq. \ref{minimised_conditions_2_main} (SimRFs+ Geometry Minimises the Truncated RF-Conformity Objective)}\label{app:simp_plus_min}
Here, we show that the SimRFs+ geometrical coupling mechanism (Eq. \ref{minimised_conditions_2_main}) minimises the truncated approximation to the RF-conformity $\tilde{\rho}(\boldsymbol{x},\boldsymbol{y})$ (Eq. \ref{eq:truncated_obj}). Writing a Lagrangian using the truncated sum and differentiating, it is straightforward to find that
\begin{equation}
\sum_{j\neq i} \left( \frac{v^{2}}{4 \Gamma(1+\frac{d}{2})}(\boldsymbol{w}_i + \boldsymbol{w}_j) \right) - \lambda_i \boldsymbol{w}_i = 0 \hspace{5ex} i=1,...,d  
\end{equation}
with the Lagrange multipliers $\lambda_i$ fixed by the (known) normalisations of $\boldsymbol{w}_i$. Should such a geometry exist, this will be solved by
\begin{equation} \label{eq:opp_dir_appendix_2}
\boldsymbol{w}_i \propto -\sum_{j\neq i} \boldsymbol{w}_j \hspace{5ex} i=1,...,d.
\end{equation}
Note that, on account of the truncation of the objective, we do \emph{not} need to make any assumptions about the vector norms or angles subtended being equal to reach this conclusion. It is straightforward to convince oneself that such a geometry always exists for any set of norms: if one norm $w_i$ exceeds the sum of all the others, Eq. \ref{eq:opp_dir_appendix_2} is trivially satisfied by arranging the vector of maximum norm to be antialigned with all the rest; if this is not the case, it is always possible to arrange the vectors such that they sum to $0$, i.e. form a closed loop. Then $\boldsymbol{w}_i = -\sum_{j\neq i} \boldsymbol{w}_j$, which satisfies Eq. \ref{eq:opp_dir_appendix_2}. We conclude that the SimRFs+ geometry 
\begin{equation} \label{eq:repeat_min_coupling}
 \boldsymbol{w}_i=-\frac{\sum_{j\neq i}  \boldsymbol{w}_j}{\|\sum_{j\neq i}  \boldsymbol{w}_j\|_2} w_i  \hspace{5ex} i=1,...,d
\end{equation}
minimises $\tilde{\rho}(\boldsymbol{x},\boldsymbol{y})$. 

We briefly note that Eq. \ref{eq:repeat_min_coupling} does not actually define one unique geometrical coupling, but empirically the iterative update scheme in Eq. \ref{simpler_updates_main} always finds a good solution when initialised in the simplex geometry. 

\subsection{Proof of Lemma \ref{iidrf_conformity} (IIDRF Conformity)}
In this appendix, we derive the probability distribution $p(w_{ij})$ over $w_{ij} = \|\boldsymbol{w}_i + \boldsymbol{w}_j\|_2$ in the case that all $\boldsymbol{w}_i$ follow independent Gaussian distributions $\mathcal{N}(0,\mathbf{I}_d)$, and use it to evaluate the corresponding IIDRF conformity $\rho(\boldsymbol{x},\boldsymbol{y})$.

In the i.i.d. case, each component of the vector $\boldsymbol{w}_i+\boldsymbol{w}_j$ is the sum of two standard normal distributions, $\mathcal{N}(0,1)$. This gives another normal distribution with twice the variance, $\mathcal{N}(0,2)$, which leads simply to the generalised $\chi_d$ distribution
\begin{equation}
     p(w_{ij}) = \frac{w_{ij}^{d-1} e^{-w_{ij}^2/4}}{2^{d-1} \Gamma(\frac{d}{2})}. 
\end{equation}
Considering the definition of $\rho(\boldsymbol{x},\boldsymbol{y})$ in Eq. \ref{eq:def_rf_conformity}, it is straightforward to calculate
\begin{equation}
\rho \left(\boldsymbol{x},\boldsymbol{y}\right) =\Gamma(\frac{d}{2}) \int_0^\infty \mathrm{d}w \frac{w^{d-1} e^{-\frac{w^2}{4}}}{2^{d-1} \Gamma(\frac{d}{2}) } \sum_{k=0}^\infty \frac{v^{2k} w^{2k}}{2^{2k} k! \Gamma(k+\frac{d}{2})} =\sum_{k=0}^\infty \frac{v^{2k}}{k!} = e^{v^2},
\end{equation}
as reported in the main text. We used the fact that all $w_{ij}$ follow the same distribution and suppressed the $ij$ subscripts for notational clarity. To perform the integral over $w$, we used the identity $\int_{w=0}^\infty \mathrm{d}w w^{2z-1} e^{-\frac{w^2}{2}} = 2^{z-1} \Gamma(z)$ . \qed

This result is obtained more quickly by noting that, following the notation in Sec. \ref{app:conformity}, $\rho(\boldsymbol{x},\boldsymbol{y}) = \frac{1}{m(m-1)}\sum_i\sum_{i\neq j}\mathbb{E}[e^{(\boldsymbol{w}_i + \boldsymbol{w}_j)^{\top} \boldsymbol{v}}] = \mathbb{E}[e^{\boldsymbol{w}_1^{\top} \boldsymbol{v}}]\mathbb{E}[e^{\boldsymbol{w}_2^{\top} \boldsymbol{v}}]$. We used the fact that $\boldsymbol{w}_i$ and $\boldsymbol{w}_j$ are independent and that all $\boldsymbol{w}_{ij}$ follow the same distribution (then choosing $i=1$ and $j=2$ wlg). We have already seen that $\mathbb{E}[e^{\boldsymbol{w}_1^{\top} \boldsymbol{v}}]=e^\frac{v^2}{2}$ (in fact the condition for unbiased estimation of $\widehat{K}$), which immediately yields $\rho(\boldsymbol{x},\boldsymbol{y})=e^{v^2}$. But the approach using $p(w_{ij})$ is a good warmup for the theorems that follow and will permit a more unified account.  

\subsection{Proof of Lemma \ref{angle_pdf} (PDF for Vectors Subtending $\theta$)} \label{app:angle_pdf_proof}
In this appendix, we derive the form of Eq. \ref{eq:theta_proby_dist}, the probability distribution of $w_{ij} = \|\boldsymbol{w}_i + \boldsymbol{w}_j\|_2$ if $\boldsymbol{w}_{i},\boldsymbol{w}_{j} \in \mathbb{R}^d$ are marginally Gaussian vectors conditioned to subtend a fixed angle $\theta$. Later, the special cases of $\theta=\frac{\pi}{2}$ (orthogonal) and $\theta = \arccos(-\frac{1}{d-1})$ (simplex) will be of particular interest.

Clearly $w^2 = w_i^2 +w_j^2 + 2w_i w_j \cos \theta$, with weight magnitudes $w_{i,j} \sim \chi_d$ (we have suppressed the $ij$ subscript, replacing $w_{ij}$ by $w$, to minimise notational clutter). Diagonalising the quadratic form, we see that a constant $w$ surface will trace out an ellipse in $(w_i,w_j)$ space with semi-major (-minor) axis lengths $\frac{w}{\sqrt{1 \pm \cos(\theta)}}$. Now
\begin{equation}
    p(w < w') = \int_\mathcal{A} p_\chi(w_i) p_\chi(w_j) \text{d}w_i \text{d}w_j
\end{equation}
where $p_\chi$ denotes the $\chi_d$ distribution obeyed by $w_{i,j}$ and $\mathcal{A}$ denotes the area in the positive quadrant bounded by an ellipse of constant $w=w'$ (recall that $w_{i,j}\geq0$ since these are vector magnitudes). Expressing this in polar coordinates,
\begin{equation}
p(w<w') = \int_{\phi =0}^{\pi/2} \text{d}\phi \int_{r=0}^{\frac{w'}{\sqrt{1+\sin(2\phi)\cos(\theta)}}} \text{d}r r p_\chi(r \cos \phi) p_\chi (r \sin \phi).
\end{equation}
Differentiating wrt $w'$ to get the pdf, 
\begin{equation}
\begin{multlined}
    p(w) = \int_{\phi =0}^{\pi/2} \text{d}\phi \frac{w}{\sqrt{1+\sin(2\phi)\cos(\theta)}} p_\chi( \frac{w  \cos \phi}{\sqrt{1+\sin(2\phi)\cos(\theta)}}) p_\chi (\frac{w  \sin \phi}{1+\sin(2\phi)\cos(\theta)})
    \\ = \frac{w^{2d-1}}{2^{d-2} \Gamma(\frac{d}{2})^2} \int_{\phi =0}^{\pi/2} \text{d}\phi ( \sin \phi \cos \phi)^{d-1} \frac{e^{-\frac{w^2}{2(1 + \sin 2 \phi \cos \theta)}}}{(1 + \sin 2 \phi \cos \theta)^{d}}, 
    \label{polarintegralinproof}
\end{multlined}
\end{equation}
as reported in Eq. \ref{eq:theta_proby_dist} of the main text. \qed

As an aside, it is instructive to set $\theta=\pi/2$ and inspect the form of $p(w)$. Doing so, we arrive at the integral
\begin{equation}
\begin{multlined}
    p(w) = \frac{w^{2d-1}}{2^{d-2} \Gamma(\frac{d}{2})^2} \int_{\phi =0}^{\pi/2} \text{d}\phi ( \sin \phi \cos \phi)^{d-1} e^{-\frac{w^2}{2}} = \frac{w^{2d-1}}{2^{2d-2} \Gamma(\frac{d}{2})^2} \int_{\phi =0}^{\pi} \text{d}\phi ( \sin \phi )^{d-1} e^{-\frac{w^2}{2}}
    \\ = \sqrt{\pi} \frac{w^{2d-1}}{2^{2d-2} \Gamma(\frac{d}{2}) \Gamma(\frac{d}{2}+\frac{1}{2})} e^{-\frac{w^2}{2}}.
    \label{reducestochi2d}
\end{multlined}
\end{equation}
Recalling the Legendre duplication formula, $\sqrt{\pi} \Gamma(2z) = 2^{2z-1} \Gamma(z)\Gamma(z+\frac{1}{2})$, this 
reduces to
\begin{equation} \label{eq:chi_2d_dist}
p(w)=\frac{w^{2d-1} e^{-w^2/2}}{2^{d-1} \Gamma(d)}.
\end{equation}
This is nothing other than the $\chi$-distribution with $2d$ degrees of freedom. This makes intuitive sense because, since $\boldsymbol{w}_i$ and $\boldsymbol{w_j}$ are orthogonal, it follows that $w^2 = w_i^2 + w_j^2$. Now $w_{i,j}$ follow $\chi_d$ distributions (square root of sum of squares of $d$ standard normal variates), so $w$ must be a square root of sum of squares of $2d$ standard normal variates -- that is, a $\chi_{2d}$ distribution.

\subsection{Proof of Theorem \ref{prf_gap_1_main} (ORF and SimRF Conformity Closed Forms)}
Here, we derive the RF-conformities $\rho(\boldsymbol{x},\boldsymbol{y})$ of the ORF and SimRF variants. 

Recall the form of $\rho(\boldsymbol{x},\boldsymbol{y})$, defined in Eq. \ref{eq:def_rf_conformity} and reproduced here for convenience: 
 \begin{equation} 
 \rho(\boldsymbol{x}, \boldsymbol{y})=
\frac{\Gamma(\frac{d}{2})}{m(m-1)} \sum_{i} \sum_{j\neq i} \mathbb{E}_{w_{ij}}\left( \sum_{k=0}^\infty \frac{v^{2k} w_{ij} ^{2k}}{2^{2k} k! \Gamma(k+\frac{d}{2})} \right).
\end{equation}
Use the probability distribution for two marginally Gaussian weights conditioned to subtend an angle $\theta$, 
\begin{equation}
   p(w_{ij}) = \frac{w_{ij}^{2d-1}}{2^{d-2} \Gamma(\frac{d}{2})^2} \int_{\phi =0}^{\pi/2} \text{d}\phi ( \sin \phi \cos \phi)^{d-1} \frac{e^{-\frac{w_{ij}^2}{2(1 + \sin 2 \phi \cos \theta)}}}{(1 + \sin 2 \phi \cos \theta)^{d}}, 
\end{equation}
where $w_{ij}=\|\boldsymbol{w}_i^2+\boldsymbol{w}_j^2\|_2$ with $i \neq j$ (see Lemma \ref{angle_pdf} and the accompanying proof in Sec. \ref{app:angle_pdf_proof}). Since all $w_{ij}$ follow the same distribution, the sums give a multiplicative factor of $m(m-1)$ that cancels with the denominator. Now we have
 \begin{equation} 
 \rho_\theta(\boldsymbol{x}, \boldsymbol{y})= \sum_{k=0}^\infty \frac{v^{2k}}{2^{2k}k! 2^{d-2} \Gamma(\frac{d}{2}) \Gamma(k+\frac{d}{2})} \int_{w=0}^\infty\mathrm{d}w \int_{\phi=0}^\frac{\pi}{2} \mathrm{d}\phi w^{2k+2d-1}(\sin\phi\cos\phi)^{d-1} \frac{e^{-\frac{w^2}{2(1+\sin2\phi\cos\theta)}}}{(1+\sin2\phi \cos\theta)^d}.
\end{equation}
Changing variables $w \to w\sqrt{1+\sin2\phi\cos\theta}$ and doing the integral over $w$,
 \begin{equation} 
 \rho_\theta(\boldsymbol{x}, \boldsymbol{y})= \sum_{k=0}^\infty \frac{v^{2k}\Gamma(k+d)}{2^{k}k! 2^{d-2} \Gamma(\frac{d}{2}) \Gamma(k+\frac{d}{2})} \int_{\phi=0}^\frac{\pi}{2}\mathrm{d}\phi (\sin2\phi)^{d-1} (1+\sin2\phi \cos\theta)^k.
\end{equation}
Finally, changing variables $\phi \to \frac{\phi}{2}$ and rearranging, we arrive at
\begin{equation}
    \rho_\theta(\boldsymbol{x},\boldsymbol{y}) = \frac{1}{2^{d-1}\Gamma(\frac{d}{2})} \int_0^{\pi}\text{d}\phi ( \sin \phi )^{d-1} \\ \cdot \sum_{k=0}^\infty \frac{v^{2k}(1 + \sin\phi\cos\theta)^k}{2^{k} k! \Gamma(k+\frac{d}{2})} \Gamma(k+d)
\end{equation}
as reported in Eq. \ref{eq:theta_conformity} of the main text.

Now we substitute in the values of $\theta$ corresponding to the particular cases of ORFs and SimRFs. 

\textbf{1) ORFs:} $\cos \theta = 0$

Note that 
\begin{equation}
\frac{1}{\Gamma(\frac{d}{2})} \int_{\phi=0}^\pi \mathrm{d}\phi (\sin\phi)^{d-1} = \sqrt{\pi} \frac{\Gamma(\frac{d}{2})}{\Gamma(\frac{d}{2}+\frac{1}{2}) \Gamma(\frac{d}{2})} = 2^{d-1} \frac{\Gamma(\frac{d}{2})}{\Gamma(d)}
\end{equation}
where we used the identity $\int_0^\pi \mathrm{d}x \sin^dx = \frac{\sqrt{\pi} \Gamma(\frac{d}{2}+\frac{1}{2})}{\Gamma(\frac{d}{2}+1)}$ and the Legendre duplication formula. It follows immediately that
\begin{equation}
\rho_\textrm{ORF}(\boldsymbol{x},\boldsymbol{y}) =\frac{\Gamma(\frac{d}{2})}{\Gamma(d)} \sum_{k=0}^\infty \frac{v^{2k}}{2^k k!} \frac{\Gamma(k+d)}{\Gamma(k+\frac{d}{2})}.
\end{equation}
We could have obtained this more directly using the $\chi_{2d}$ distribution (see discussion at the end of Sec. \ref{app:angle_pdf_proof}), but leaving $\theta$ unspecified for as long as possible permits a more direct comparison with SimRFs.

\textbf{2) SimRFs:} $\cos \theta = -\frac{1}{d-1}$

Carrying out the binomial expansion,
\begin{equation}
\begin{multlined}
\int_{0}^\pi \mathrm{d}\phi (\sin\phi)^{d-1} \left(1-\frac{\sin\phi}{d-1}\right)^k = \sum_{p=0}^k \frac{k!}{(k-p)!p!}\left(-\frac{1}{d-1}\right)^p \int_{0}^\pi \mathrm{d}\phi (\sin \phi) ^{d+p-1}
\\ = \sum_{p=0}^k \frac{k!}{(k-p)!p!}\left(-\frac{1}{d-1}\right)^p \sqrt{\pi} \frac{\Gamma(\frac{d+p}{2})}{\Gamma(\frac{d+p+1}{2})}.
\end{multlined}
\end{equation}
Substituting this in, we immediately arrive at
\begin{equation}
 \rho_\mathrm{SimRF}(\boldsymbol{x},\boldsymbol{y}) =\frac{\sqrt{\pi}}{\Gamma(\frac{d}{2}) 2^{d-1}}\sum_{k=0}^\infty \frac{\Gamma(k+d)}{\Gamma(k+\frac{d}{2})}  \frac{v^{2k}}{2^k} \sum_{p=0}^k \left(-\frac{1}{d-1}\right)^p \frac{\Gamma(\frac{d+p}{2})}{\Gamma(\frac{d+p+1}{2})} \frac{1}{(k-p)! p!}
\end{equation}
which we have seen is smaller than $\rho_\mathrm{ORF}(\boldsymbol{x},\boldsymbol{y})$. \qed

\subsection{Proof of Corollary \ref{corr:orth_gap} (ORFs Always Outperform IIDRFs)}
Here we derive an analytic expression for the orthogonality gap (difference in kernel estimator MSE between the IIDRF and ORF mechanisms) and show that it is positive everywhere.

From Eq. \ref{mse_and_conformity}, we immediately have that
\begin{equation}
\Delta \mathrm{MSE}(\widehat{K}(\mathbf{x},\mathbf{y})) =e^{-2(x^2+y^2)} \frac{m-1}{m} \left( \rho_\mathrm{IIDRF}(\boldsymbol{x},\boldsymbol{y}) - \rho_\mathrm{ORF}(\boldsymbol{x},\boldsymbol{y}) \right).
\end{equation}
Inserting the respective RF-conformities from Eqs. \ref{eq:iidrf_conformity} and \ref{eq:orf_conformity},
\begin{equation}
\begin{multlined}
\Delta \mathrm{MSE}(\widehat{K}(\mathbf{x},\mathbf{y})) = e^{-2(x^2+y^2)} \frac{m-1}{m} \left(e^{v^2}-\frac{\Gamma(\frac{d}{2})}{\Gamma(d)} \sum_{k=0}^\infty \frac{v^{2k}}{2^k k!} \frac{\Gamma(k+d)}{\Gamma(k+\frac{d}{2})} \right)
\\ = e^{-2x^2-2y^2} \frac{m-1}{m}
\sum_{k=0}^\infty \frac{v^{2k}}{k!} \left(1-\frac{(k+d-1)!}{(d-1)!} \frac{(d-2)!!}{(2k+d-2)!!}\right),
\end{multlined}
\end{equation}
where $!!$ denotes the double factorial. We can write the term in parentheses as 
\begin{equation}
1-\frac{d}{d}\cdot\frac{d+1}{d+2}\cdot ...\cdot \frac{d+k-2}{d+2(k-2)}\cdot\frac{d+k-1}{d+2(k-1)}>0
\end{equation}
so the series expansion is positive. It follows that the kernel estimator MSE with ORFs is upper bounded but that of IIDRFs. \qed

\subsection{Proof of Theorem \ref{thm:rff_gap} (RFF Orthogonality Gap)}\label{app:rff_results}
Here, we demonstrate how, with minor modifications, many of the stated results for PRFs can be translated to RFFs. 

Recall the RFF definition, stated in Eq. \ref{rffdef2} of the main text and reproduced here for convenience.
\begin{equation}
\begin{multlined}
    \phi_{\mathrm{RFF}}(\boldsymbol{z}) \overset{\mathrm{def}}{=} \sqrt{\frac{1}{m}} (\odot_{i=1}^{m}[\sin(\boldsymbol{w}_i^{\top} \boldsymbol{z}),\cos(\boldsymbol{w}_i^{\top} \boldsymbol{z}) ])^{\top}.
    \end{multlined}
\end{equation}
Now we have that 
\begin{equation}
    \widehat{K} =\phi(\boldsymbol{x})^{\top} \phi(\boldsymbol{y}) \\= \frac{1}{m}\sum_{i=1}^m \cos \boldsymbol{w}_i^\top (\boldsymbol{x}-\boldsymbol{y}) = \frac{1}{m} \sum_{i=1}^m a_i
\end{equation}
where we defined 
\begin{equation}
a_i \overset{\mathrm{def}}{=} \cos(\boldsymbol{w}_i^{\top}\boldsymbol{z})
\end{equation}
and let $\boldsymbol{z}=\boldsymbol{x}-\boldsymbol{y}$. It is straightforward to show that $\widehat{K}$ is an unbiased estimator of the Gaussian kernel $e^{-\frac{\|\boldsymbol{x}-\boldsymbol{y}]|_2^2}{2}}$ when we sample $\boldsymbol{w}_i \sim \mathcal{N}(0,\mathbf{I}_d)$; that is, $\mathbb{E}[a_i] = e^{-\frac{z^2}{2}}$. After some work, we also have that
\begin{equation}
    \text{MSE}(\widehat{K})= \frac{1}{m}\left (\frac{(1-e^{-z^2})^2}{2} +  (m-1)(\frac{1}{m(m-1)}\sum_i\sum_{i\neq j}\mathbb{E}[a_ia_j]-e^{-z^2}) \right).
\label{prf_variance_2}
\end{equation}
The object of interest (which is precisely the analogue of $\rho(\boldsymbol{x},\boldsymbol{y})$ but for RFFs) is $\sum_i\sum_{i\neq j}\mathbb{E}[a_ia_j] = \sum_i\sum_{i\neq j}\mathbb{E}[\cos \boldsymbol{w}_i^\top\boldsymbol{z}\cos \boldsymbol{w}_j^\top\boldsymbol{z}]$. It will vary depending on any geometrical coupling scheme employed. 

From elementary trigonometry, $\cos \boldsymbol{w}_i^\top\boldsymbol{z}\cos \boldsymbol{w}_j^\top\boldsymbol{z} = \frac{1}{2}(\cos\left((\boldsymbol{w}_i+\boldsymbol{w}_j)^\top \boldsymbol{z}\right)+\cos\left((\boldsymbol{w}_i-\boldsymbol{w}_j)^\top \boldsymbol{z}\right)$. It is also simple to convince oneself that, when the random vectors $\boldsymbol{w}_i$ and $\boldsymbol{w}_j$ are (a) i.i.d. or (b) conditioned to be orthogonal, the distributions of the two random variables $\boldsymbol{w}_i+\boldsymbol{w}_j$ and $\boldsymbol{w}_i-\boldsymbol{w}_j$ are identical. As such, we can just consider the single random variable $\boldsymbol{w}_{ij} = \boldsymbol{w}_i + \boldsymbol{w}_j$ wlg. Then we have that
\begin{equation}
\begin{multlined}
\mathbb{E}[\cos \boldsymbol{w}_{ij}^{\top} \boldsymbol{z}] = \int_{\mathbb{R}_d} p(\boldsymbol{w}_{ij}) e^{-i \boldsymbol{w}_{ij}^{\top}\boldsymbol{z}} \text{d}^d\boldsymbol{w}_{ij}
\\ = \Gamma(d/2) 2^{\frac{d}{2}-1} \int_0^\infty \text{d}w_{ij} p(w_{ij}) (w_{ij}z)^{1-\frac{d}{2}} J_{\frac{d}{2}-1} (w_{ij}z)
\end{multlined}
\end{equation}
where we have used that the probability distribution $p(\boldsymbol{w}_{ij})$ is real regardless of whether the random vectors are i.i.d. or orthogonal, and written the expression as a Hankel transform \cite{arizona_hankel}. Note that we do \emph{not} consider the simplex coupling case, where the random variables $\boldsymbol{w}_i+\boldsymbol{w}_j$ and $\boldsymbol{w}_i-\boldsymbol{w}_j$ will follow different distributions.

Carefully comparing with Eq. \ref{eq:prf_hankel} in Sec. \ref{app:conformity}, we observe that \textbf{the expression is identical to the PRF case, but instead taking $v \to -iz$}. This means that we can obtain all the previously stated IIDRF and ORF results in the RFF setting with minimal extra work. For instance, inspecting Eq. \ref{eq:prf_orth_gap}, we can immediately state the RFF orthogonality gap (difference in kernel estimator MSE between IIDRFs and ORFs) reported in Theorem \ref{thm:rff_gap}:
\begin{equation}
\Delta \text{{MSE}}(\widehat{K}(\mathbf{x},\mathbf{y})) = \frac{m-1}{m} \left( e^{-z^2} - \frac{\Gamma(d/2)}{\Gamma(d)} \sum_{k=0}^\infty \frac{(-z^2)^k}{2^k k!} \frac{\Gamma(k+d)}{\Gamma (k+d/2)} \right).
\end{equation}
(Note that we also dropped the exponential prefactor $e^{-2x^2-2y^2}$, originating from the definition of PRFs (\ref{eq:def_prf}) where it is needed to keep kernel estimation unbiased). \qed

\subsection{Proof of Corollary \ref{corr:asymptotic_rff_form} (RFF Asymptotic MSE Ratio)}
Here we derive Eq. \ref{eq:asymptotic_ratio}, the ratio of ORF to IIDRF kernel estimator MSEs in the $d \to \infty$ limit. This was first reported in \cite{yu2016orthogonal}, and is included here to show consistency with our more general (finite $d$) closed forms.

Considering the discussion in Sec. \ref{app:rff_results}, it is straightforward to reason that the ratio of MSEs is given by
\begin{equation} \label{eq:appendix_ratio}
 \frac{\text{MSE}_\text{ORF}}{\text{MSE}_\text{IIDRF}} = 1 + \frac{2(m-1)}{(1+e^{-z^2})^2} (\mathbb{E}_\text{ort}(a_1a_2) - e^{-z^2})
\end{equation}
where $a_i=e^{-i\boldsymbol{w}_i^\top\boldsymbol{z}}$ and the expectation is being taken over the random variable $w_{12}=\|\boldsymbol{w}_1 + \boldsymbol{w}_2\|_2$, with $\boldsymbol{w}_{1,2}$ conditioned to be orthogonal (see e.g. Eq. \ref{eq:chi_2d_dist} for the appropriate probability distribution). From the discussion in Sec. \ref{app:rff_results}, the term in parentheses on the right can be written as the series expansion
\begin{equation}
\sum_{k=0}^\infty \frac{(-z^2)^k}{k!} \left( \frac{1}{2^k} \frac{\Gamma(k+d) \Gamma(\frac{d}{2})}{\Gamma(k+\frac{d}{2}) \Gamma(d)} -1\right ).
\end{equation}
Recalling Stirling's formula for the asymptotic form of the Gamma function, 
\begin{equation}
\lim_{x \to \infty} \Gamma(x+1) = \sqrt{2\pi x} e^{-x} x^x,
\end{equation}
we can rewrite term
\begin{equation}
\begin{multlined}
\lim_{d\to\infty}\frac{1}{2^k} \frac{\Gamma(k+d) \Gamma(\frac{d}{2})}{\Gamma(k+\frac{d}{2}) \Gamma(d)} = \frac{1}{2^k} \sqrt{ \frac{(k+d-1)(\frac{d}{2}-1)}{(k+\frac{d}{2}-1)(d-1)}} \frac{(k+d-1)^{m+d-1} (\frac{d}{2}-1)^{\frac{d}{2}-1}}{(k+\frac{d}{2}-1)^{k+\frac{d}{2}-1} (d-1)^{d-1}}
\\ = \sqrt{ \frac{(k+d-1)(\frac{d}{2}-1)}{(k+\frac{d}{2}-1)(d-1)}}  \cdot \frac{1}{2^k} \left ( \frac{k+d-1}{k+\frac{d}{2}-1} \right)^k \cdot \frac{\left( 1 + \frac{k}{d-1} \right )^{d-1} }{ \left ( 1 + \frac{k}{\frac{d}{2}-1} \right )^ {\frac{d}{2}-1} }.
\end{multlined}
\end{equation}
We can Taylor expand each of the constituent components, 
\begin{equation} 
\sqrt{ \frac{(k+d-1)(\frac{d}{2}-1)}{(k+\frac{d}{2}-1)(d-1)}} \simeq 1 - \frac{k}{2d}
\end{equation}
\begin{equation}
\frac{1}{2^k} \left ( \frac{k+d-1}{k+\frac{d}{2}-1} \right)^k  \simeq 1 - \frac{k(k-1)}{d} 
\end{equation}
\begin{equation}
\frac{\left( 1 + \frac{k}{d-1} \right )^{d-1} }{ \left ( 1 + \frac{k}{\frac{d}{2}-1} \right )^ {\frac{d}{2}-1} } \simeq 1 + \frac{k^2}{2d}
\end{equation}
which combine to yield
\begin{equation}
\frac{1}{2^k} \frac{\Gamma(k+d) \Gamma(\frac{d}{2})}{\Gamma(k+\frac{d}{2}) \Gamma(d)} \simeq 1 - \frac{k(k-1)}{2d}.
\end{equation}
It follows that
\begin{equation}
    \sum_{k=0}^\infty \frac{(-z^2)^k}{k!} \left( \frac{1}{2^k} \frac{\Gamma(k+d) \Gamma(\frac{d}{2})}{\Gamma(k+\frac{d}{2}) \Gamma(d)} -1\right ) = -\sum_{k=0}^\infty \frac{(-z^2)^k}{k!} \frac{k(k-1)}{2d}  = -\frac{z^4}{2d} e^{-z^2}.
\end{equation}
Putting this into Eq. \ref{eq:appendix_ratio},
\begin{equation}
\frac{\text{MSE}(\widehat{K}_\text{ORF})}{\text{MSE}(\widehat{K}_\text{IIDRF})} = 1 - (m-1)\left(\frac{e^{-z^2} z^4}{d(1-e^{-z^2})^2} + \mathcal{O}\left(\frac{1}{d^2}\right)\right), 
\end{equation}
as reported in Eq. \ref{eq:asymptotic_ratio} and \cite{yu2016orthogonal}. Importantly, the negative sign of the subleading term means that, in the RFF setting, ORFs will always outperform IIDRFs when $d \to \infty$. \qed

\section{Experimental Details}\label{exp_appendix}
In this appendix, we provide further experimental details to supplement the discussion in Sec. \ref{experiments}.

\subsection{Choosing $\sigma$} \label{sigmachoice}
\begin{figure}[H]
\centering
  \includegraphics[width = 0.8\linewidth]{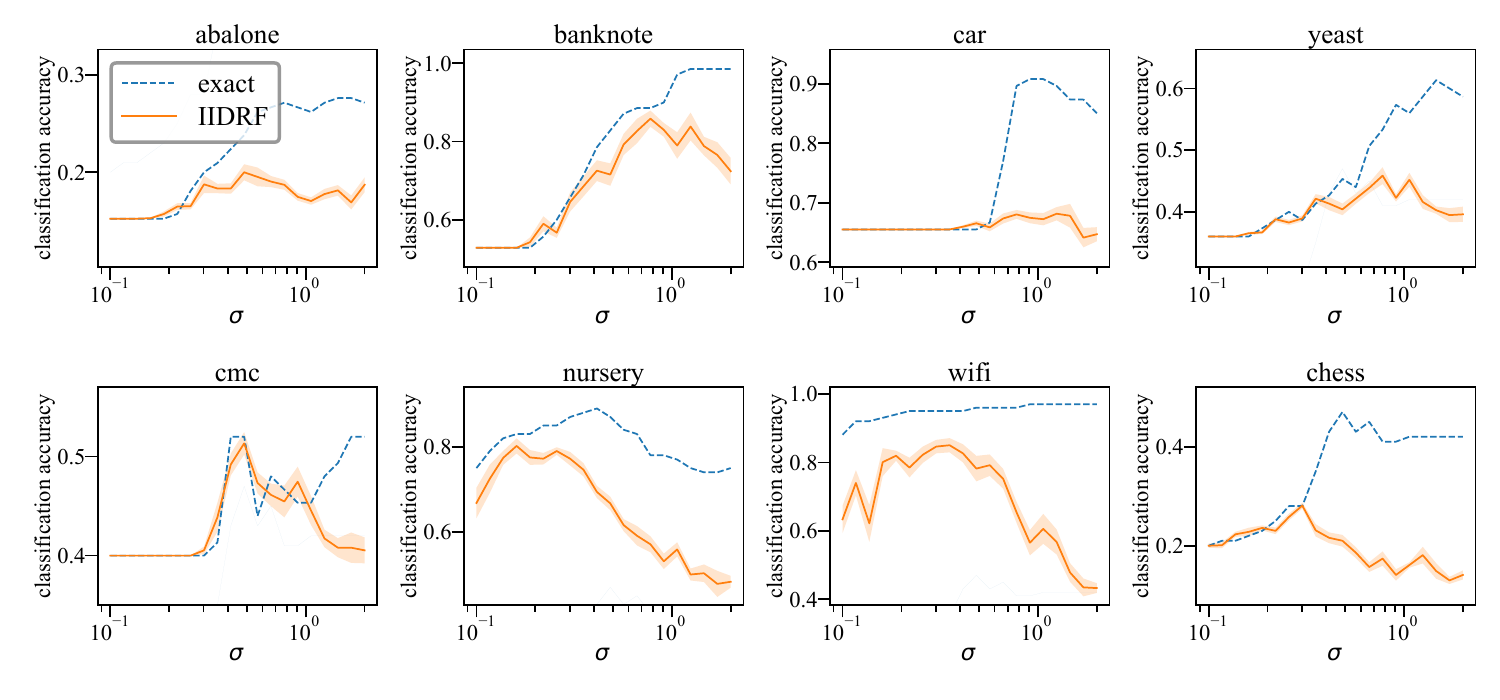}
  \caption{Plots showing classification accuracy vs the data rescaling factor $\sigma$ for each of the nonparametric classification validation datasets, including both the exact (dotted blue) and IIDRF (solid orange) kernels. They are used for tuning the $\sigma$ hyperparameter. The RFs are of dimensionality $m=10d$, with $d$ the data dimensionality, and the shaded region gives one standard deviation on the estimate of the mean classification accuracy over $N=10$ samples. During this coarse $\sigma$ search phase, the large $\mathrm{nursery}$ and $\mathrm{chess}$ datasets are restricted to $1000$ training examples and $100$ test examples for speed.}
  \label{tunegraph}
\end{figure}

Here we elaborate on Sec \ref{nonparam_class}, where we report tuning the hyperparameter $\sigma$ with a  validation dataset. In particular, given some fixed dataset $(\boldsymbol{x},\boldsymbol{y})$ and a Gaussian kernel $K(\boldsymbol{x},\boldsymbol{x}') = e^{-\frac{\|\boldsymbol{x}-\boldsymbol{x}'\|_2^2}{2}}$, we apply the scalar transformation $(\boldsymbol{x},\boldsymbol{y}) \to (\sigma \boldsymbol{x},\boldsymbol{y})$ with $\sigma \in \mathbb{R}^+$ to optimise the IIDRF performance. There are two (potentially competing) factors to consider:
\begin{enumerate}
\item $\sigma$ implicitly controls the smoothness of the the kernel $K$ that we are approximating. Multiplying the data by $\sigma$ is equivalent to rescaling the kernel characteristic lengthscale by $\frac{1}{\sigma}$, i.e. taking $K(\boldsymbol{x},\boldsymbol{x}') = e^{-\frac{\|\boldsymbol{x}-\boldsymbol{x}'\|_2^2}{2}} \to e^{-\frac{\|\boldsymbol{x}-\boldsymbol{x}'\|_2^2}{2/\sigma^2}}$. This will change the classifier accuracy even when using the exact kernel.
\item The kernel estimator variance has some dependence on the data $(\boldsymbol{x},\boldsymbol{x}')$ -- consider e.g. any of the results in Sec. \ref{sec:theory}, or Fig. \ref{var_comp_fig}. Roughly speaking, PRFs tend to perform worse at large $\sigma$ (equivalent to a sharply varying kernel).  
\end{enumerate}
In order to navigate a possible tradeoff between these factors and pick a suitable value for $\sigma$, we tune by running a coarse search optimising classifier accuracy on a validation set. This is sensible because we are specifically interested in comparing the performance of SimRFs, ORFs and IIDRFs in settings where kernel approximation with random features is already effective. Fig. \ref{tunegraph} shows the results; we choose the value of $\sigma$ at which the i.i.d. PRF classification accuracy (orange solid line) peaks. As we have suggested, this does \emph{not} generically coincide with where the exact kernel performance (blue dotted line) peaks. 

There is also a broader question about the relationship between the kernel estimator MSE and the performance in downstream tasks. The fact that SimRFs consistently outperform ORFs and IIDRFs in a variety of nonparametric classification tasks and when used to approximate the attention module in Performers confirms that lower variance on kernel estimates often helps performance in applications. But making rigorous mathematical statements about the effect of estimator variance -- especially, why its importance differs between tasks -- is complicated and is left as an open question. 

\subsection{Fast SimRFs: Further Discussion and Experimental Results}\label{app:fastsimrfs}
In this appendix, we provide more detailed discussion of fast SimRFs (Sec. \ref{sec:fastsimrfs}) and demonstrate a simple implementation on the nonparametric classification task. Experiments will be closely related to those described in Sec. \ref{nonparam_class} so readers are advised to review this section first.

Recall the definition of the simplex block,
\begin{equation}
\mathbf{W}_\textrm{simp} = \mathbf{DSR},
\end{equation}
where $\mathbf{D} \in \mathbb{R}^{d \times d} = \mathrm{diag}(w_i)$ with $w_i$ sampled from a $\chi_d$-distribution. $\mathbf{R} \in \mathbb{R}^{d \times d}$ is a random orthogonal matrix drawn from Haar measure on $\mathrm{O}(d)$, the group of orthogonal matrices in $\mathbb{R}^{d\times d}$. The rows $\boldsymbol{s}_i$ of the simplex projection matrix $\mathbf{S} \in \mathbb{R}^{d \times d}$ are given by the simplex unit vectors, defined in Eq. \ref{eq:simplex_vectors} and reproduced below for convenience. 
\begin{equation}
\label{appeq:simplex_vectors}
\boldsymbol{s}_i = \begin{cases}
     \sqrt{\frac{d}{d-1}} \textbf{e}_i - \frac{\sqrt{d}+1}{(d-1)^{3/2}} (1,...,1,0)^{\top}  & \text{for}\ 1 \leq i < d \\
      \frac{1}{\sqrt{d-1}}(1,1,...,1,0)^{\top} & \text{for}\ i = d. \\
    \end{cases}
\end{equation}
Recall further that, with fast SimRFs, we replace the matrix $\mathbf{R}$ by an orthogonal proxy $\widetilde{\mathbf{R}}$ that is only approximately sampled from Haar measure, but which supports fast matrix-vector multiplication.   

\subsubsection{$\mathbf{S}$ Supports Fast Matrix-Vector Multiplication}\label{app:fastsimsalg}
We begin by showing that the matrix $\mathbf{S}$ supports fast matrix-vector multiplication, as stated in the main body. The following simple algorithm of time-complexity $\mathcal{O}(d)$ calculates $\mathbf{S}\boldsymbol{x}$, with $\boldsymbol{x}\in\mathbb{R}^d$.
\begin{algorithm}[H]
   \caption{Fast matrix-vector multiplication with $\mathbf{S}$}
   \label{alg:fastmult}
\begin{algorithmic}
   \STATE {\bfseries Input:} object vector $\boldsymbol{x}\in\mathbb{R}^d$ with components $x_i, i=1,...,d$
   \STATE {\bfseries Output:} `simplex projection' vectors $\boldsymbol{y} = \mathbf{S}\boldsymbol{x}\in\mathbb{R}^d$
   \STATE {\bfseries Main:}
   \STATE $y_d = \frac{1}{\sqrt{d-1}} \sum_{i=1}^{d-1} x_i$
   \vspace{1mm}
   \FOR{$i=1$ {\bfseries to} $d-1$}
   \vspace{1mm}
   \STATE $y_i = \sqrt{\frac{d}{d-1}} x_i - \frac{\sqrt{d}+1}{d-1} y_d$
   \vspace{1mm}
   \ENDFOR
\end{algorithmic}
\end{algorithm}
We note that, in downstream applications such as the SimRFs-Performer, the time taken by this extra simplex projection (whether using the fast implementation or not) is typically dwarfed by other computational requirements. It is rarely observable. That said, the extra time cost compared to ORFs is technically nonzero and constitutes the only real weakness of SimRFs.

\subsubsection{Implementation using $\mathbf{HD}$-Product Matrices}\label{app:fastsimsexp}
To demonstrate one possible implementation, we use the so-called $\mathbf{HD}$-product matrices, formed by multiplication of $k\in\mathbb{N}$ $\mathbf{HD}$ blocks, 
\begin{equation}
\widetilde{\mathbf{R}} = \prod_{i=1}^k \mathbf{H}\mathbf{D}_i^{(\mathcal{R})}. 
\end{equation}
Here, $\mathbf{H}$ is the normalised Hadamard matrix, defined by the recursive relation
\begin{equation}
\mathbf{H}_1 = (1), \hspace{5mm} \mathbf{H}_i = \frac{1}{\sqrt{2}} \left( \begin{matrix}
\mathbf{H}_{i-1} & \mathbf{H}_{i-1}\\
\mathbf{H}_{i-1} & -\mathbf{H}_{i-1}
\end{matrix}\right) \hspace{2mm}\textrm{for   } i>1,
\end{equation}
and $\mathbf{D}_i^{(\mathcal{R})} = \textrm{diag}{(d_i)}$ with $d_i \sim \textrm{Unif}(\{\pm1\})$, i.i.d. Rademacher random variables. $\mathbf{HD}$-blocks have previously received attention for dimensionality reduction \cite{chazelle}, locally-sensitive hashing methods \cite{bojarski2017structured} and kernel approximation \cite{choromanski2017unreasonable}, where they exhibit good computational and statistical properties. Importantly, given some vector $\boldsymbol{x} \in \mathbb{R}^d$, the matrix-vector product $\mathbf{H}\boldsymbol{x}$ can be computed in time $\mathcal{O}(d \log d)$ via the fast Walsh-Hadamard transform.

We report the results of the nonparametric classification tasks described in Sec. \ref{nonparam_class}, now inserting $\mathbf{HD}$-product matrices with $k=3$ in place of $\mathbf{R}$. With $m=d$ random features, we observe that using fast SimRFs and fast ORFs does not substantially change the accuracy of nonparametric classification. Results are frequently identical to the regular case. 

\begin{table}[H]
\caption{
Classification accuracies from kernel regression with IIDRFs, ORFs and SimRFs, where we include both regular and fast implementations in the latter two cases. Replacing the random orthogonal matrix $\mathbf{R}$ (sampled from Haar measure) with a structured $\mathbf{HD}$-product does not change the accuracy of nonparametric classification.}
\label{tab:fastsims_table}\vspace{-0mm}
\begin{center}
\begin{small}
\begin{tabular}{l|c|cc|cr}
\toprule 
 & \multicolumn{5}{c}{Classification accuracy}  \\ \cmidrule{2-6}
Dataset&  IIDRFs & \multicolumn{2}{c|}{ORFs}  & \multicolumn{2}{c}{SimRFs}\\
&  & Regular & Fast & Regular & Fast\\
\midrule
$\mathrm{abalone}$    & $0.1432 \pm 0.0003$ & $0.1445 \pm 0.0003$ & $0.1447 \pm 0.0003$ &$0.1455 \pm 0.0003$ & $0.1462 \pm 0.0003$  \\
$\mathrm{banknote}$ & $0.6441 \pm 0.0024$ &$0.6612 \pm 0.0025$ & $0.6596 \pm 0.0024$ & $0.7196\pm0.0019$ &$0.7296 \pm 0.0017$ \\
$\mathrm{car}$    & $0.6768 \pm 0.0006$ &$0.6788 \pm 0.0006$ & $0.6784 \pm 0.0006$ &$0.6797\pm0.0006$&$0.6800 \pm 0.0006$\\
$\mathrm{yeast}$    & $0.3187 \pm 0.0006$ & $0.3193\pm0.0006$ & $0.3171 \pm 0.0006$ &$0.3187\pm0.0006$ & $0.3195 \pm 0.0006$   \\
$\mathrm{cmc}$     & $0.4088 \pm 0.0009$ &$0.4149\pm0.0009$ & $0.4159 \pm 0.0009$ & $0.4206\pm0.0008$&$0.4222 \pm 0.0008$ \\
$\mathrm{nursery}$      & $0.5870 \pm 0.0013$ &$0.6213\pm0.0019$ & $0.6193 \pm 0.0019$ & $0.7030\pm0.0008$& $0.7037 \pm 0.0008$ \\
$\mathrm{wifi}$      &$0.4914 \pm 0.0026$ &$0.5224\pm0.0025$ & $0.5310 \pm 0.0024$ & $0.6509\pm0.0027$ &$0.6533 \pm 0.0027$       \\
$\mathrm{chess}$   & $0.2011 \pm 0.0002$ &$0.2017\pm0.0002$ & $0.2016 \pm 0.0002$ &$0.2021\pm0.0002$ & $0.2021 \pm 0.0002$ \\
\bottomrule
\end{tabular}
\end{small}
\end{center}
\vskip -0.1in
\end{table}

Note that Hadamard matrices are defined such that $\mathbf{H}\in\mathbb{R}^{d\times d}$ with $d=2^i, i \in \mathbb{N}$ a non-negative integer. Given data of some arbitrary dimensionality, we have to `pad' each object vector $\boldsymbol{x}$ with $0$s such that its length is a power of $2$. This accounts for the small discrepancies between the results reported in Table \ref{tab:fastsims_table} and accuracies in Fig. \ref{nonparam_results}, where vectors are not padded so $m=d$ is smaller. Results in each column of the table are for the same effective $d$ so the regular and fast mechanisms can be safely compared.


\end{document}